\documentclass{article}

\usepackage[preprint]{neurips_2026}

\usepackage{comment}            
\usepackage[utf8]{inputenc} 
\usepackage[T1]{fontenc}    
\usepackage{hyperref}       
\usepackage{url}            
\usepackage{booktabs}       
\usepackage{amsfonts}       
\usepackage{amssymb}        
\usepackage{nicefrac}       
\usepackage{microtype}      
\usepackage{xcolor}         
\usepackage{enumitem}       
\usepackage{amsmath}        
\usepackage{amsthm}         
\usepackage{graphicx}       
\usepackage{multirow}       
\usepackage{array}          
\graphicspath{{figs/}}      

\title{Action Emergence from Streaming Intent}

\author{%
  Pengfei Jing$^{1,2*}$\quad
  Victor Shea-Jay Huang$^{1,3}$\quad
  Hengtong Lu$^{1,2}$ \\[0.3em]
  \bf Jifeng Dai$^{2}$\quad
  Yan Xie$^{1}$\quad
  Benjin Zhu$^{1,2*\dagger}$ \\[0.3em]
  \small $^{1}$Li Auto\quad
  $^{2}$Tsinghua University\quad
  $^{3}$CUHK \\[0.1em]
  \small $^{*}$Equal contribution\quad
  $^{\dagger}$Corresponding author\\
  \normalfont\small
  Project page:\, \url{https://mind-omni.github.io/}
}

\begin{document}

\maketitle

\begin{abstract}
We formalize \emph{action emergence} as a target capability for end-to-end
autonomous driving: the ability to generate physically feasible, semantically
appropriate, and safety-compliant actions in arbitrary, long-tail traffic scenes
through scene-conditioned reasoning rather than retrieval or interpolation of
learned scene-action mappings.
We show that previous paradigms cannot deliver action emergence:
autoregressive trajectory decoders collapse the inherently multimodal future
into a single averaged output, while diffusion and flow-matching generators
express multimodality but are not steerable by reasoned intent.
We propose \textbf{Streaming Intent} as a concrete way to approach action
emergence: a mechanism that makes driving intent (i)~\emph{semantically
streamed} through a continuous chain-of-thought that causally derives the
intent from scene understanding, and (ii)~\emph{temporally streamed} across
clips so that intent commitments remain coherent along the driving horizon.
We realize Streaming Intent in a VLA model we call \textbf{SI}
(\emph{Streaming Intent}).
SI autoregressively decodes a four-step chain-of-thought and emits an intent
token; the decoded intent then drives classifier-free guidance (CFG) on a
flow-matching action head, requiring only two denoising steps to generate the
final trajectory.
On the Waymo End-to-End benchmark, SI achieves competitive aggregate
performance, with an RFS score of 7.96 on the validation set and 7.74 on the
test set.
Beyond aggregate metrics, the model demonstrates~--~to our knowledge for the
first time in a fully end-to-end VLA~--~\emph{intent-faithful controllability}:
for a fixed scene, varying the intent class at inference yields qualitatively
distinct yet consistently high-quality plans, arising purely from data-driven
learning without any pre-built trajectory bank or hand-coded post-hoc selector.
\end{abstract}

\section{Introduction}
\label{sec:intro}

Despite rapid progress on aggregate planning benchmarks, end-to-end autonomous
driving systems remain brittle on the long tail: rare junction geometries,
unprotected turns, dense merging, and ambiguous yielding scenarios continue to
drive the bulk of human-takeover events in deployed fleets.
We argue that a core missing capability is \emph{action emergence}: the ability
to produce physically feasible, semantically appropriate, and safety-compliant
actions in arbitrary scenes through on-the-fly reasoning over perceptual and
contextual inputs, rather than through retrieval or interpolation of previously
learned scene-action mappings.
Central to this capability is \textbf{driving intent}: a discrete high-level
commitment~--~yield, merge, turn, cruise~--~that mediates between scene
understanding and trajectory generation.
Without an explicit intent representation, an agent has no structured basis for
committing to one future among equally plausible alternatives; the long-tail
failure modes of current systems are, in large part, failures of intent.%
\footnote{Unlike scale-driven LLM emergence~\citep{wei2022emergent}, our action emergence is a scale-independent, application-level behavior realizable by any scene-conditioned reasoning architecture.}

\begin{figure}[t]
  \centering
  \includegraphics[width=\linewidth]{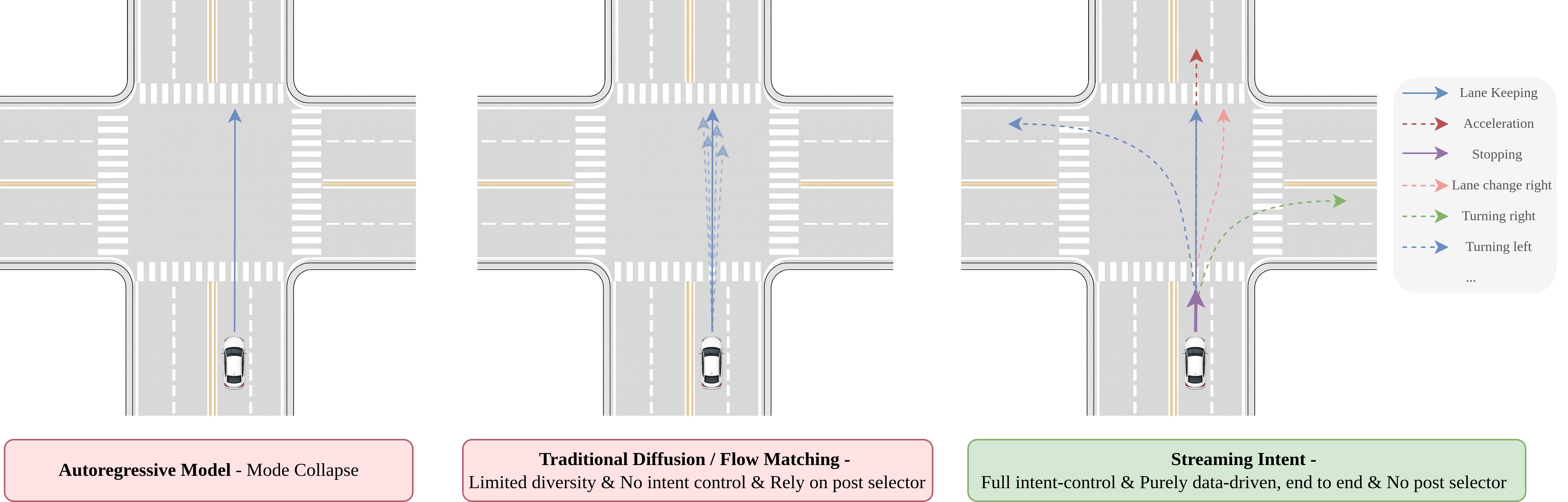}
  \caption{\textbf{Trajectory diversity under ambiguous intent.}
  Given the same intersection scene, AR models collapse to a single averaged
  future, diffusion/FM models sample a narrow prior-dominated trajectory bundle,
  whereas \textbf{SI} produces intent-faithful trajectories.}
  \label{fig:diversity_compare}
\end{figure}

\textbf{Prior trajectory generators cannot deliver action emergence.}
Prior end-to-end trajectory generators fall into two families that each fail to provide the intent commitment required
for action emergence (\autoref{fig:diversity_compare}).
\textbf{(i)~Autoregressive (AR) trajectory models}~\citep{zhou2025autovla,rowe2025poutine,luo2025adathinkdrive,chen2026curiousvla}
decode future waypoints token by token, which tends to collapse the inherently
multimodal future into a single averaged trajectory.
At an ambiguous junction, this can produce a physically unrealizable compromise
between ``turn left'', ``continue straight'', and ``yield''.
\textbf{(ii)~Diffusion and flow-matching (FM) trajectory models}~\citep{ho2020denoising,lipman2023flowmatching,liao2025diffusiondrive,zheng2025diffusion,xing2025goalflow,xu2025wamflow,li2025recogdrive}
can represent multimodal trajectory distributions, but without an explicit
conditioning signal indicating \emph{which} mode to commit to, sampling is
dominated by the data prior and remains weakly steerable by reasoned intent.
Neither family delivers \emph{intent-faithful controllability}~--~the property
that a specific, explicitly reasoned intent determines the output
trajectory~--~which we identify as a necessary condition for action emergence:
without a mechanism that binds reasoned intent to executed trajectory, an agent
cannot commit to a plan in scenes where multiple intents are plausible.

\textbf{Our approach: SI and Streaming Intent.}
We propose \textbf{SI} (\emph{Streaming Intent}), a VLA model
built around the concept of
\textbf{Streaming Intent}~--~a mechanism that approaches action emergence by
making intent both semantically grounded in scene reasoning and temporally
coherent across the driving horizon.
SI comprises three tightly integrated components.

\textbf{(a)~Single-backbone language--action alignment.}
SI operates on a single shared transformer backbone~\citep{vaswani2017attention} that serves both the
autoregressive (AR) language decoder and the flow-matching (FM) action head~\citep{lipman2023flowmatching,liu2022flow}.
The AR branch decodes the chain-of-thought and emits an intent token; the
decoded intent class then drives classifier-free guidance (CFG)~\citep{ho2022classifierfree} on the FM head.
Because both objectives train the \emph{same} representation end-to-end,
language reasoning and trajectory denoising are structurally coupled:
language--action alignment is achieved \emph{by construction} rather than by
an auxiliary loss or a post-hoc bridging module.

\textbf{(b)~Intent-driven CFG for intent-faithful trajectory generation.}
The decoded intent directly conditions the FM denoising process via CFG~\citep{ho2022classifierfree},
steering the generated trajectory toward the committed maneuver rather than
defaulting to the statistically dominant mode.
At inference, supplying different intent classes to the same trained model
yields geometrically and behaviorally distinct trajectories for the same
scene~--~constituting the \emph{intent-faithful controllability} that prior
models lack and that we identify as a necessary condition for action emergence.

\textbf{(c)~Streaming Intent: semantic and temporal continuity.}
Streaming Intent makes intent continuous along two dimensions.
\emph{Semantic streaming}: intent is not predicted as an isolated label, but
emerges from a four-step chain-of-thought~\citep{wei2022chainofthought}
(\textsc{Perceive} $\rightarrow$ \textsc{Predict} $\rightarrow$
\textsc{Judge} $\rightarrow$ \textsc{Plan}) before being emitted as the intent
token; dense CoT annotation makes intent a scene-grounded intermediate
representation rather than an independent classifier output.
\emph{Temporal streaming}: the current clip's intent token and LLM hidden state
are compressed into a compact memory token and carried to the next clip, so
each intent prediction is conditioned on accumulated episode history without
recomputing the full backbone.
Together, these two forms of streaming make intent a dynamically evolving,
causally grounded, and temporally coherent commitment that bridges VLA reasoning
and FM action generation toward action emergence.

\textbf{Results.}
On the Waymo End-to-End benchmark~\citep{xu2025wode2e}, SI achieves an
RFS of $\mathbf{7.96}$ on the validation split, and an
RFS of $\mathbf{7.74}$ on the test split
(\autoref{subsec:planning_quality}, \autoref{tab:planning},
\autoref{tab:planning_test}).
Beyond aggregate numbers, SI demonstrates two capabilities that, to our
knowledge, no prior end-to-end VLA has shown from a single trained model:
(i)~\emph{action emergence} on long-tail scenes, where SI's
intent-conditioned trajectories span the plausible action repertoire while
strong single-mode baselines collapse onto the dominant
forward-cruising mode (\autoref{subsec:action_emergence},
\autoref{fig:action_emergence}); and (ii)~\emph{intent-faithful
controllability}, where varying the intent for a fixed scene at inference
yields geometrically distinct yet uniformly high-quality plans that align
with the human-rated RFS alternatives rather than exhibit random variance
(\autoref{subsec:multi_intent_quality}, \autoref{fig:multi_intent_quality}).
Critically, this controllable diversity arises from a single
end-to-end-trained model on structurally aligned language--action
representations~--~it is not stitched from a pre-built trajectory bank, nor
selected by a hand-tuned post-hoc preference module, as in prior
multi-trajectory approaches~\citep{chai2020multipath,phanminh2020covernet,chen2024vadv2,sun2026sparsedrivev2,gao2026rad2}.

\medskip
\noindent
In summary, our contributions are:
\begin{itemize}[leftmargin=1.2em,itemsep=2pt,topsep=2pt]
  \item We \textbf{formalize action emergence} as the application-level capability that end-to-end autonomous driving systems
        should aspire to, and identify driving intent as the structural
        ingredient whose absence prevents current VA and VLA models from
        approaching it.
  \item We propose \textbf{Streaming Intent} and instantiate it in
        \textbf{SI}: a single-backbone VLA in which AR-decoded intent
        drives CFG on a shared flow-matching action head, with intent grounded
        through four-step CoT (semantic streaming) and carried across clips via
        a prev-intent memory token (temporal streaming)~--~together a
        concrete realization of action emergence.
  \item On Waymo End-to-End~\citep{xu2025wode2e}, SI achieves competitive
        aggregate performance with an RFS score of 7.96 on the validation set
        and 7.74 on the test set, while demonstrating~--~to our knowledge the
        first time in a fully end-to-end VLA~--~\emph{intent-faithful
        controllability} arising purely from data-driven learning, without any
        pre-built trajectory bank or hand-coded trajectory selector.
\end{itemize}

\section{Method}
\label{sec:method}

\begin{figure}[t]
  \centering
  \noindent\makebox[\linewidth][c]{%
    \includegraphics[width=1.2\linewidth]{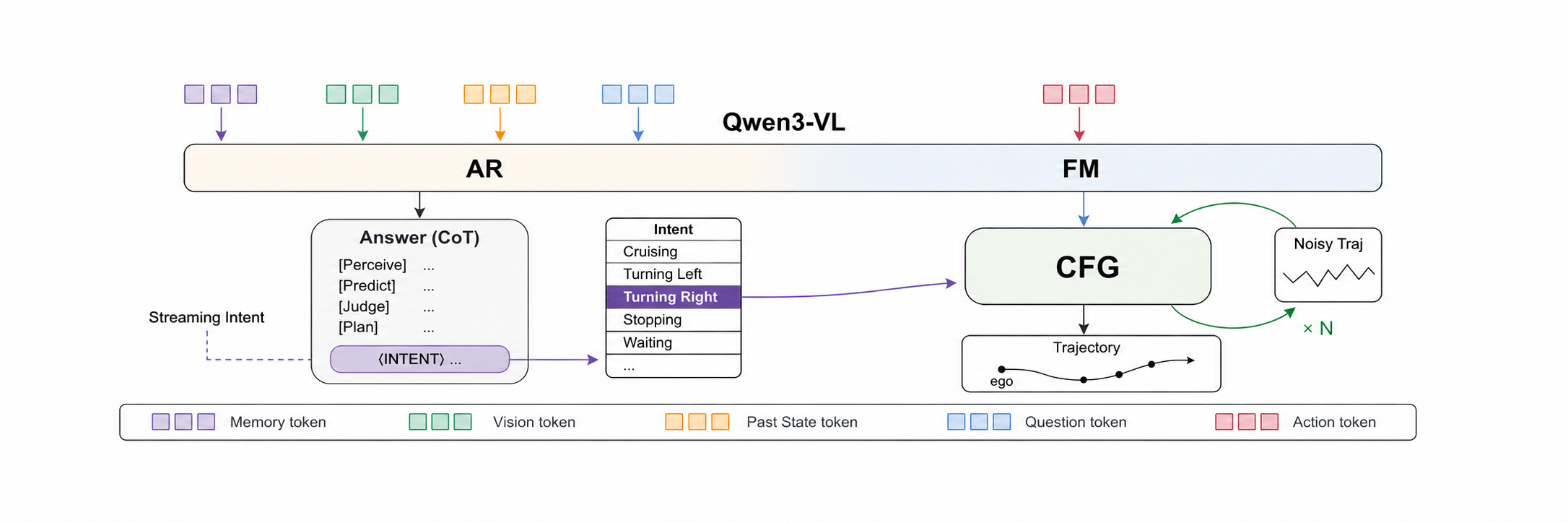}%
  }
  \caption{\textbf{SI architecture.}
  A single shared Qwen3-VL backbone jointly supports AR CoT/intent decoding,
  FM intent-guided trajectory denoising, with streaming intent.}
  \label{fig:si_arch}
\end{figure}

\subsection{Overview}
\label{subsec:overview}

\autoref{fig:si_arch} shows the full SI architecture: a single
Qwen3~\citep{yang2025qwen3} backbone is shared by the AR language branch and
the FM action branch in one forward pass.  The AR branch consumes memory,
vision, past-state, and CoT-question tokens to decode a four-step
chain-of-thought
($\textsc{Perceive} \rightarrow \textsc{Predict} \rightarrow
\textsc{Judge} \rightarrow \textsc{Plan}$) ending in
$\langle\textsc{INTENT}\rangle k\langle/\textsc{INTENT}\rangle$, while the FM
branch consumes noisy action tokens to denoise the future trajectory under
intent-conditioned flow matching.  A lightweight \emph{intent bridge} parses
the AR span into one of 20 intent classes $k \in [0, 19]$, providing the
symbolic handoff from reasoning to action; because both branches share the same
backbone weights and causal attention, language and action are aligned by
construction rather than through an auxiliary loss.

\subsection{Streaming Intent to CFG Trajectory Generation}
\label{subsec:streaming_intent}

This subsection traces the path from the AR-decoded intent span to
the final trajectory on the right of~\autoref{fig:si_arch}.
Let $K=20$ be the number of driving-intent classes, $H$ the backbone
hidden size, $T$ the trajectory chunk length, and $D$ the per-step
action dimension.

\textbf{Intent bridge.}
For each clip, the AR half emits an answer ending in a discrete span
$\langle\textsc{INTENT}\rangle\,\textit{name}\,\langle/\textsc{INTENT}\rangle$.
A lightweight bridge regex-extracts the span and looks \textit{name}
up against a fixed 20-class taxonomy, yielding
$k \in \{0, \ldots, K-1\}$; parse failures (missing span or unknown
name) fall back to a default class, so the FM head always receives a
well-formed signal.
The parsed $k$ is embedded by a small two-layer MLP over a learned
table of $K{+}1$ rows (the extra row at index $K$ is the
\emph{unconditional} slot), producing $e(k) \in \mathbb{R}^H$; this
vector is added, broadcast along the action-chunk axis, as a
token-wise bias to the flow-matching action--time stream inside the
shared backbone, so that cond and uncond forwards differ \emph{only}
in this single bias term.

\textbf{Flow-matching objective and CFG dropout.}
The trajectory is generated by rectified flow on the action chunk
$x_0 \in \mathbb{R}^{T \times D}$.
At training time we sample $t \sim \mathrm{Beta}(1.5, 1.0)$ and
Gaussian noise $\varepsilon \sim \mathcal{N}(0, I)$, form the
interpolant $x_t = t\,\varepsilon + (1-t)\,x_0$, and train the
backbone-conditioned velocity head $v_\theta$ to regress
$u_t = \varepsilon - x_0$ with an MSE loss.
Classifier-free guidance is enabled by \emph{CFG dropout}: with
probability $p_{\text{drop}}$ the intent index is replaced by the
uncond index $K$, and samples flagged as pseudo-labeled are
\emph{always} replaced. A single network thus fits both the
intent-conditional and unconditional velocity fields.

\textbf{CFG-guided rectified-flow sampling.}
At inference we evaluate two forward passes per denoising step, one
with the parsed intent $k$ and one with the uncond index $K$, and
combine the resulting velocity fields:
\begin{equation}
  v_t \;=\; v_\theta(x_t, t;\, e(K))
         \;+\; w\,\big[\,v_\theta(x_t, t;\, e(k))
                      \,-\, v_\theta(x_t, t;\, e(K))\,\big],
\label{eq:cfg}
\end{equation}
and the clean trajectory is recovered by Euler integration of
rectified flow from $t{=}1$ to $t{=}0$ in $N$ uniform steps,
\begin{equation}
  x_{t + \Delta t} \;=\; x_t + \Delta t\, v_t,
  \qquad
  \Delta t = -\tfrac{1}{N}.
\label{eq:euler}
\end{equation}
\emph{Why this matters.} Because cond and uncond share the \emph{same}
backbone and differ only in the bias $e(\cdot)$, the guidance term
$v_\theta(\cdot;\,e(k)) - v_\theta(\cdot;\,e(K))$ isolates exactly the
trajectory direction that the CoT-derived intent adds on top of the
unconditional scene prior, and $w$ amplifies that direction.
Intent therefore acts as a \emph{continuous steering dimension of a
single data-driven generative model}: one pass of end-to-end training
over paired $\langle$scene, CoT, intent, trajectory$\rangle$ data
yields a network in which simply swapping the intent index at
inference produces qualitatively different yet equally plausible
plans for the same scene.
No pre-built trajectory bank, no mode-wise decoder ensemble, and no
hand-tuned post-hoc mode selector are introduced at any stage.
The intent-faithful controllability that prior AR and
unconditional-FM trajectory generators cannot deliver
(\autoref{sec:intro}) emerges, in SI, as a direct consequence
of classifier-free guidance applied to a language-conditioned FM
head -- purely from data.

\textbf{Halving inference cost via CFG distillation.}
Eq.~\eqref{eq:cfg} incurs a $2\times$ inference cost: every denoising
step runs two backbone forwards.
After the main training has converged, we distil this two-pass
operator into a \emph{single-pass} student embedder
\begin{equation*}
  e_{\text{dist}}(k) \;=\; \bar{e}(k) \;+\; \mathrm{MLP}\big(\bar e(k)\big),
  \qquad
  \bar e(k) \;\;\text{warm-started to}\;\;
  w\,e(k) - (w{-}1)\,e(K),
\end{equation*}
whose warm-start row $\bar e(k)$ is the closed-form linear target of
Eq.~\eqref{eq:cfg} (the CFG-effective vector an idealized linear
backbone would consume), and whose residual MLP is trained to absorb
the nonlinear correction the actual Qwen3-VL backbone introduces.
With the rest of the network frozen (only ${\sim}21\,\mathrm{M}$
student parameters are trainable), the student is trained purely by
\emph{velocity-level} matching of the teacher's CFG output at every
denoising step, evaluated on the teacher's own $x_t$ trajectory so
teacher and student stay aligned at each step.
At deployment the FM head is routed through $e_{\text{dist}}$
instead of $e(\cdot)$: the intent-steered trajectory is recovered in
a \emph{single} backbone forward per denoising step, \textbf{halving
action-head inference cost} with negligible trajectory degradation.

\textbf{Streaming Intent: semantic and temporal.}
We now make precise what ``Streaming Intent'' names in the title of
this subsection -- it refers to two complementary flows, both
visible as rightward motion in~\autoref{fig:si_arch}.
\emph{(i)~Semantic streaming, within a clip.}
Intent is not independently predicted; it is the \emph{causal
continuation} of the four-step CoT
($\textsc{Perceive} \rightarrow \textsc{Predict} \rightarrow
\textsc{Judge} \rightarrow \textsc{Plan} \rightarrow
\langle\textsc{INTENT}\rangle$), decoded autoregressively on the
shared backbone so that every earlier step attends to the video,
memory, and past-state tokens and every later step attends back to
all preceding tokens.
The emitted $k$ is therefore the \emph{conclusion} of scene
understanding, which is what makes it a trustworthy driver of
Eq.~\eqref{eq:cfg}.
\emph{(ii)~Temporal streaming, across clips.}
The committed $k_t$ at clip~$t$ is additionally fed forward at the
symbolic level: a small prev-intent table $E^{\text{prev}} \in
\mathbb{R}^{(K+1)\times H}$, kept deliberately \emph{separate} from
the CFG embedder $e(\cdot)$ so that gradients do not interfere,
emits a single memory token $E^{\text{prev}}[k_t]$ that is prepended
to clip~$t{+}1$'s memory stream entering the AR half (an ``unknown''
row is used for the first clip, parse failures, and pseudo-labels).
Clip~$t{+}1$'s CoT is thus explicitly conditioned on the intent
committed at clip~$t$, closing the streaming loop.
Together, the semantic stream on the AR side and the symbolic memory
token across clips make intent a dynamically evolving, temporally
coherent commitment -- the Streaming Intent mechanism that bridges
language reasoning and trajectory generation in SI.

\subsection{Data Construction}
\label{subsec:data_construction}

The Streaming Intent mechanism of \autoref{subsec:streaming_intent} requires
dense per-clip supervision of three semantically linked quantities:
(i) a kinematic \emph{meta-action}, (ii) a closed-set \emph{driving intent}
grounded in scene reasoning, and (iii) auxiliary scene-understanding QAs that
anchor the VLM's perception and intent-anchored CoT texts during training.  Since no existing driving corpus
provides this triplet at scale, we build a four-stage annotation pipeline that
converts raw driving sequences into streams of fully labelled clips with
essentially zero human annotation cost.

\textbf{Stage 1: Streaming clip extraction.}
Raw sequences are uniformly downsampled to $2\,\mathrm{Hz}$ and partitioned
into streaming clips indexed by $t = 1,2,\ldots,N$.  Each clip contains a
fixed-length past observation window, a fixed-length future window of the same
vehicle, a 16-step ego past-state trace, and a 20-point BEV future trajectory
at $4\,\mathrm{Hz}$.  The $2\,\mathrm{Hz}$ sampling preserves a
driving-relevant visual horizon within the VLM context budget, while the
clip-time indexing matches the sequence on which the prev-intent memory token
of \autoref{subsec:streaming_intent} operates.

\textbf{Stage 2: Rule-based meta-action annotation.}
For each clip, a deterministic rule engine reduces future kinematics to a pair
of closed-set \emph{meta-actions}: a \emph{lateral} label
(\textsc{Keep Lane}, \textsc{Lane Change Left}, \textsc{Lane Change Right},
\dots) from sustained yaw-rate and cumulative lateral offset, and a
\emph{longitudinal} label (\textsc{Accelerate}, \textsc{Decelerate},
\textsc{Maintain Speed}, \textsc{Stop}, \dots) from the signed change in
forward speed over the future window.  This annotation is fully reproducible,
requires zero human labelling, serves as an explicit input hint to the VLM in
Stage~3, and provides the kinematic consistency check for validating the VLM's
free-form intent output.

\textbf{Stage 3: VLM-based streaming CoT for intent, with auxiliary VQA.}
For each clip, a SOTA vision-language model
(Qwen3.5-~\citep{qwen3.5}) receives the past video, future video, and Stage-2
meta-action, then emits a structured answer whose four reasoning steps
($\textsc{Perceive}\!\rightarrow\!\textsc{Predict}\!\rightarrow\!\textsc{Judge}\!\rightarrow\!\textsc{Plan}$)
run as consecutive autoregressive steps, each attending to the full video and
the running prefix of previous steps.  The final emission is a discrete intent
span
$\langle\textsc{INTENT}\rangle\,\textit{name}\,\langle/\textsc{INTENT}\rangle$,
where \textit{name} belongs to a fixed 20-class taxonomy
(e.g., \emph{go straight}, \emph{turn left}, \emph{turn right},
\emph{pull over}, \emph{park}, \dots).  This streaming CoT makes intent a
\emph{causal conclusion} of scene understanding rather than an independent
label, which SI inherits during training and replays at inference.  Every
generated intent is then checked against the Stage-2 meta-action
(e.g., \textit{turn left} requires sustained leftward yaw); inconsistent
samples are re-labelled with the rule-derived fallback class so that the FM
head never receives kinematically absurd intent supervision.  In parallel, the
same VLM produces scene-understanding question--answer pairs spanning
object-centric, spatial, temporal, motion, and common-sense categories, which
accompany the CoT supervision to ground perception during SI training.

\textbf{Stage 4: Per-clip aggregation.}
The outputs of Stages 1--3 are bundled into a unified per-clip record
$\{\text{past video}, \text{future video}, \text{past state},
\text{meta-action}, \text{CoT}, \text{intent}, \text{VQA},
\text{BEV trajectory}\}$ and concatenated along the clip-time axis into the
final streaming training set.  Each record can be used as an independent
supervised sample, while the preserved temporal order enables sequence-level
training of the streaming-memory and prev-intent pathways
(\autoref{subsec:streaming_intent}).  The pipeline is run once offline over
the training corpus; the resulting annotations are reproducible, closed-set by
construction, and carry essentially no human-labelling cost.  Full details of
the intent and meta-action taxonomies, rule-based kinematic thresholds, VLM
intent-annotation prompt, and a worked example are provided in
\autoref{app:data_construction}.

\section{Experiments}
\label{sec:exp}

\begin{table}[t]
  \centering
  \footnotesize
  \begin{minipage}[t]{0.48\linewidth}
    \centering
    \setlength{\tabcolsep}{4pt}
    \caption{\textbf{WOD-E2E val split.}}
    \label{tab:planning}
    \resizebox{!}{1.35cm}{%
    \begin{tabular}{l c c}
      \toprule
      Method & RFS & TR (\%) \\
      \midrule
      WAM-Flow(~\cite{xu2025wamflow})         & 5.17          & 14.0          \\
      Curious-VLA(~\cite{chen2026curiousvla}) & 5.81          & 30.0          \\
      AutoVLA(~\cite{zhou2025autovla})        & 6.74          & 46.0          \\
      RecogDrive(~\cite{li2025recogdrive})    & 7.40          & 58.0          \\
      RAP(~\cite{feng2025rap})                & 7.91          & \textbf{70.7} \\
      \midrule
      \textbf{SI} (ours)                      & \textbf{7.96} & 63.5          \\
      \bottomrule
    \end{tabular}%
    }
  \end{minipage}%
  \hfill
  \begin{minipage}[t]{0.48\linewidth}
    \centering
    \setlength{\tabcolsep}{4pt}
    \caption{\textbf{WOD-E2E test split.}}
    \label{tab:planning_test}
    \resizebox{!}{1.35cm}{%
    \begin{tabular}{l c c c}
      \toprule
      Method & RFS & ADE$_{3\mathrm{s}}$ & ADE$_{5\mathrm{s}}$ \\
      \midrule
      Open-LLaMA                            & 7.43          & 1.31          & 3.22          \\
      NaiveEMMA                             & 7.53          & 1.32          & 3.02          \\
      AutoVLA (~\cite{zhou2025autovla})     & 7.56          & 1.35          & 2.96          \\
      dVLM-AD (~\cite{ma2025dvlm})          & 7.63          & 1.29          & 3.02          \\
      HMVLM (~\cite{wang2025hmvlm})         & 7.74          & 1.33          & 3.07          \\
      \midrule
      \textbf{SI} (ours)                    & \textbf{7.74} & \textbf{1.24} & \textbf{2.81} \\
      \bottomrule
    \end{tabular}%
    }
  \end{minipage}
\end{table}

\subsection{Implementation}
\label{subsec:impl}

\textbf{Architecture.}
SI is built on a single \emph{Qwen3-VL-2B-Instruct}~\citep{yang2025qwen3} transformer
backbone (hidden size $H{=}1536$) that carries both the AR language
branch and the FM action branch.
The flow-matching head operates on an action chunk of length
$T{=}20$ sampled at $4\,\mathrm{Hz}$ ($5\,\mathrm{s}$ horizon,
$D{=}2$ planar coordinates) and is rolled out with $N{=}2$ Euler
steps.
The total model has $2.46\,\mathrm{B}$ parameters, of which
$2.07\,\mathrm{B}$ are trainable end-to-end ($84.5\%$); the
remainder are the Qwen3-VL visual encoder's DeepStack projectors
and tokenizer embeddings that we freeze throughout.

\textbf{Streaming Intent.}
The intent taxonomy contains $K{=}20$ driving classes plus a single
unconditional row (index $K$) used for CFG dropout and uncond
branching.
The intent embedder uses an internal dimension of $512$, trained
with CFG-dropout probability $p_{\text{drop}}{=}0.15$; at inference
we run Eq.~\eqref{eq:cfg} with a \emph{guidance scale $w{=}1.5$}.
A prototype-regression auxiliary loss (weight $0.1$) regularizes
the intent-embedder output to carry trajectory-geometric information,
and the prev-intent streaming table uses the same $p_{\text{drop}}$.
The cross-clip memory compressor produces $128$ tokens per clip
through $6$ cross-attention layers with $16$ heads, writes into a
FIFO bank of capacity $256$, and is read back with an
SE($2$) ego-motion positional encoding so that the AR half sees
geometrically aligned history at each new clip.

\textbf{Optimization.}
We train with AdamW ($\mathrm{lr}{=}1{\times}10^{-4}$, weight decay
$0.1$, $\beta_1{=}0.9$, $\beta_2{=}0.999$) under a OneCycle
schedule with warmup fraction $0.05$ and cosine anneal, for
$75$ epochs.
Gradients are clipped at norm $0.5$ and training runs in bf16
AMP on $8$ GPUs with a per-GPU batch size of $1$ \emph{sequence}
(clip groups of length $3$--$6$).
The training corpus contains $20\,745$ clips across $4\,122$
continuous subsequences drawn from the public Waymo Open Dataset
End-to-End Camera (WOD~E2E~CAM~v1.0.0).
Each training step consumes one whole subsequence; the
streaming-memory bank and the prev-intent token therefore receive
dense, within-sequence gradient across every training iteration.
Across the $75$ epochs, two-thirds of the steps are supervised
with the auxiliary VQA data and one-third with the 4-step CoT
data; the two supervision modes are \emph{randomly mixed} at the
step level so the model simultaneously retains the LLM
backbone's general VQA competence and acquires the
expert CoT reasoning that Streaming Intent relies on.
At every step both objectives update the shared backbone
jointly: an \textbf{AR loss} (teacher-forcing) supervises the
language branch, and a \textbf{flow-matching loss} supervises
the trajectory branch, so language reasoning and trajectory
denoising are co-trained on the same parameters on every
iteration.
Training and evaluation are orchestrated with the EFG deep
learning framework~\citep{zhu2023efg}, which provides the
efficient and flexible sequence-level training loop used
throughout this work.

\textbf{CFG distillation.}
After the main model converges, we run a short distillation stage
on the same training corpus: all model parameters are frozen except
the Distilled\-Intent\-Embedder ($\sim 21\,\mathrm{M}$ trainable
parameters, inner dim $3072$); the base row is warm-started to
$w\,e(k) - (w{-}1)\,e(K)$ (\autoref{subsec:streaming_intent}) and
the residual MLP is fit to minimize the per-step velocity-MSE
against the CFG teacher.
At deployment we simply replace the intent-embedder call with the
distilled version, which recovers the steered velocity field in a
single backbone forward per denoising step.

\subsection{Planning Quality}
\label{subsec:planning_quality}

We evaluate on the WOD-E2E rater-annotated splits, comprising
$438$ validation sequences and the official test leaderboard~\cite{xu2025wode2e}.
We report the official Rater Feedback Score (RFS), the
within-trust-region rate (TR) on the validation split, and the
average displacement error (ADE) at $3\,\mathrm{s}$ /
$5\,\mathrm{s}$ horizons on the test set.
SI numbers are produced by a single per-clip forward of our final
training checkpoint on the full WOD-E2E validation
(\autoref{tab:planning}) and test (\autoref{tab:planning_test})
sets; baseline numbers are reproduced from the respective works.
On the validation set, SI reaches an RFS of $\mathbf{7.96}$,
improving on the strongest prior RFS baseline
(RAP~\cite{feng2025rap}, $7.91$) by $+0.05$ absolute RFS and
outperforming every reported baseline on RFS.
RAP attains the highest TR ($70.7\%$): as an end-to-end model without
LLM-based initialization, RAP converges tightly onto the learned
trajectory distribution and therefore falls inside the trust region
with high probability.
RFS, however, is the more informative metric for our claim because it
measures closeness to \emph{human-rated high-score} trajectories and
serves as the official headline score of WOD-E2E, whereas TR only
checks whether the prediction falls inside a learned envelope around
the demonstration.
On the test set, SI reaches an RFS of $\mathbf{7.74}$ with ADE
$\mathbf{1.24}\,\mathrm{m}$ / $\mathbf{2.81}\,\mathrm{m}$ at the
$3\,\mathrm{s}$ / $5\,\mathrm{s}$ horizons, consistent with the
validation-set ranking.
Open-LLaMA and NaiveEMMA appear without citations in
\autoref{tab:planning_test} because their numbers are transcribed
directly from the WOD-E2E leaderboard and no corresponding
publication is publicly available.

\begin{figure}[t]
  \centering
  \setlength{\tabcolsep}{2pt}
  \renewcommand{\arraystretch}{1.05}
  \noindent\makebox[\linewidth][c]{%
  \begin{tabular}{@{}c@{\hspace{4pt}}c@{\hspace{4pt}}c@{}}
    \footnotesize\textbf{SI (Action Emergence)} &
    \footnotesize\textbf{RAP (Mode Collapse)} &
    \footnotesize\textbf{SI vs.\ RAP -- BEV} \\
    \includegraphics[height=0.27\linewidth,keepaspectratio]%
      {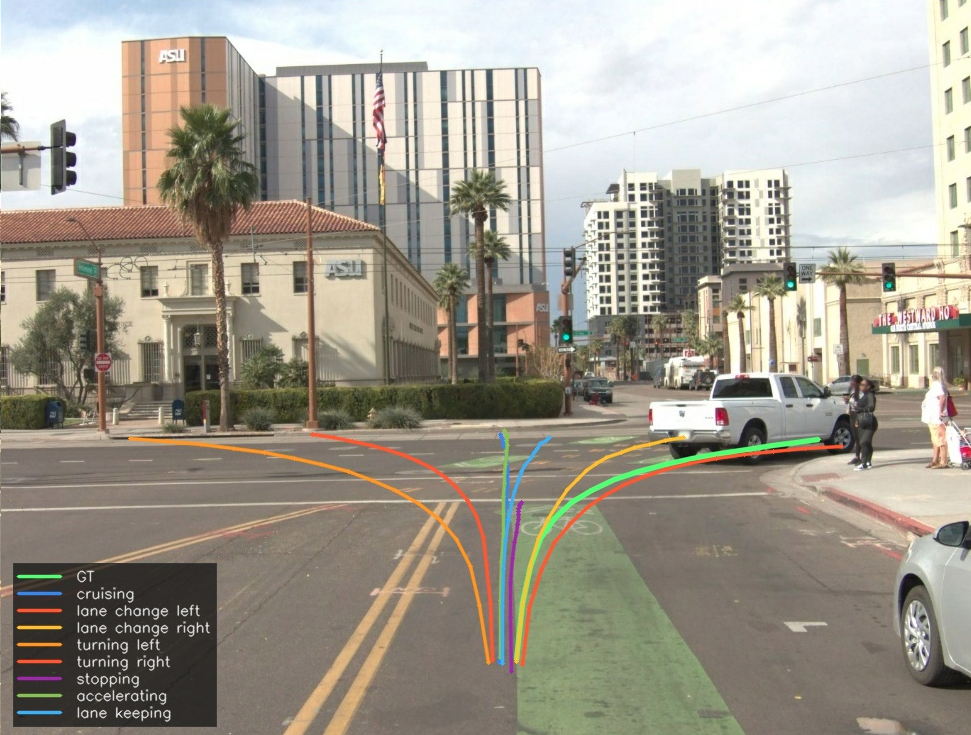} &
    \includegraphics[height=0.27\linewidth,keepaspectratio]%
      {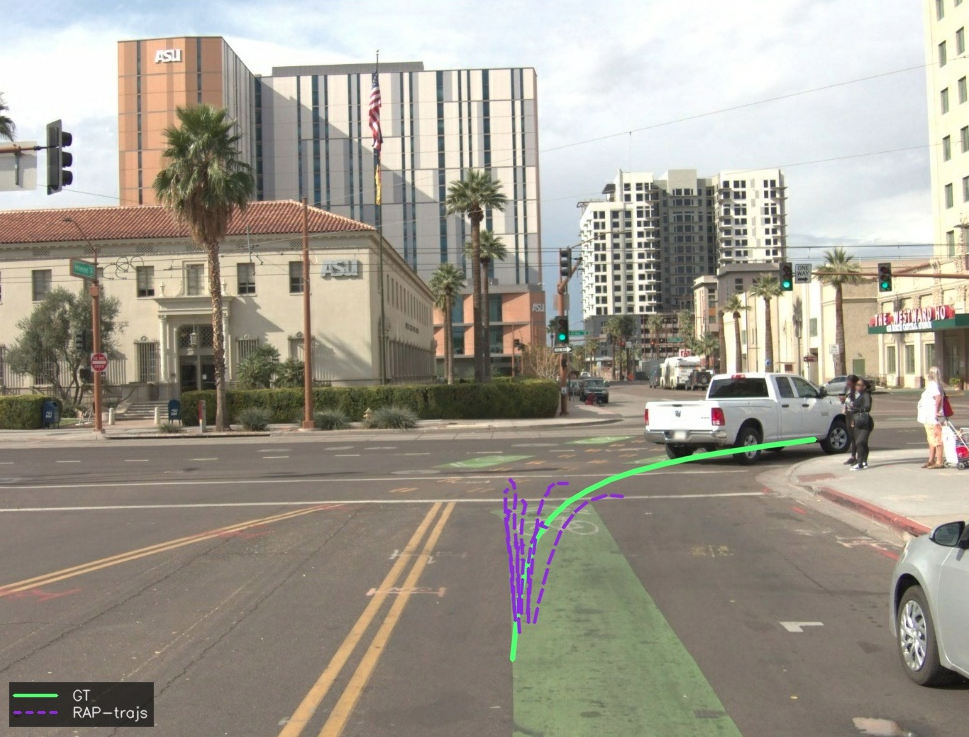} &
    \includegraphics[height=0.27\linewidth,keepaspectratio]%
      {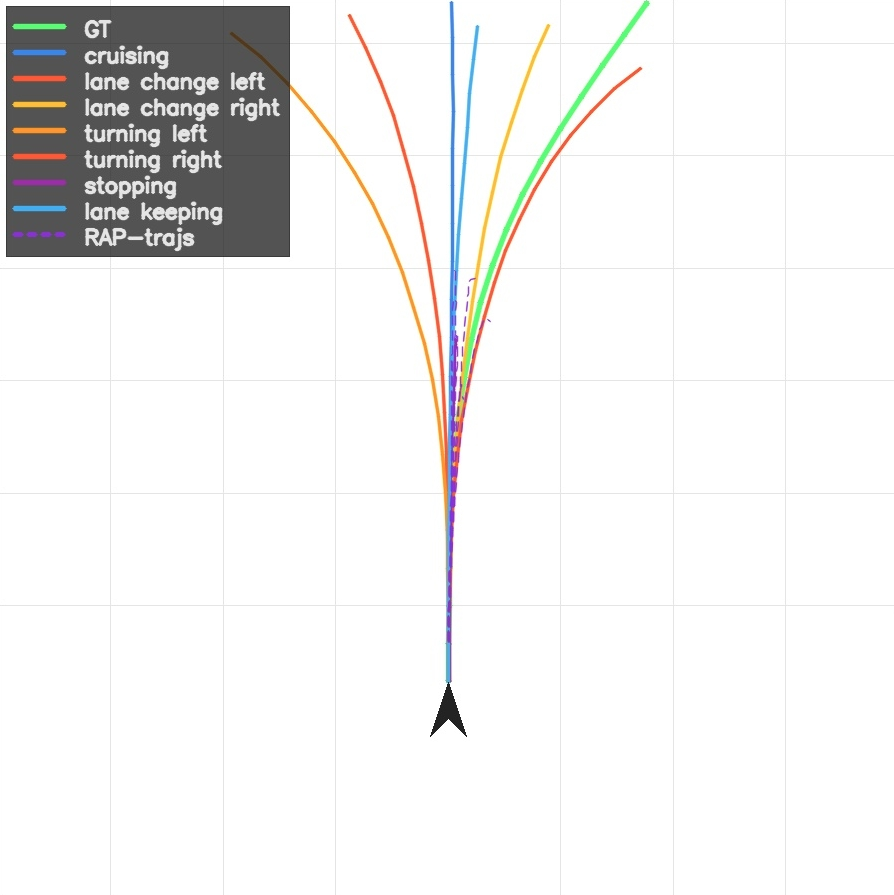} \\
    \includegraphics[height=0.27\linewidth,keepaspectratio]%
      {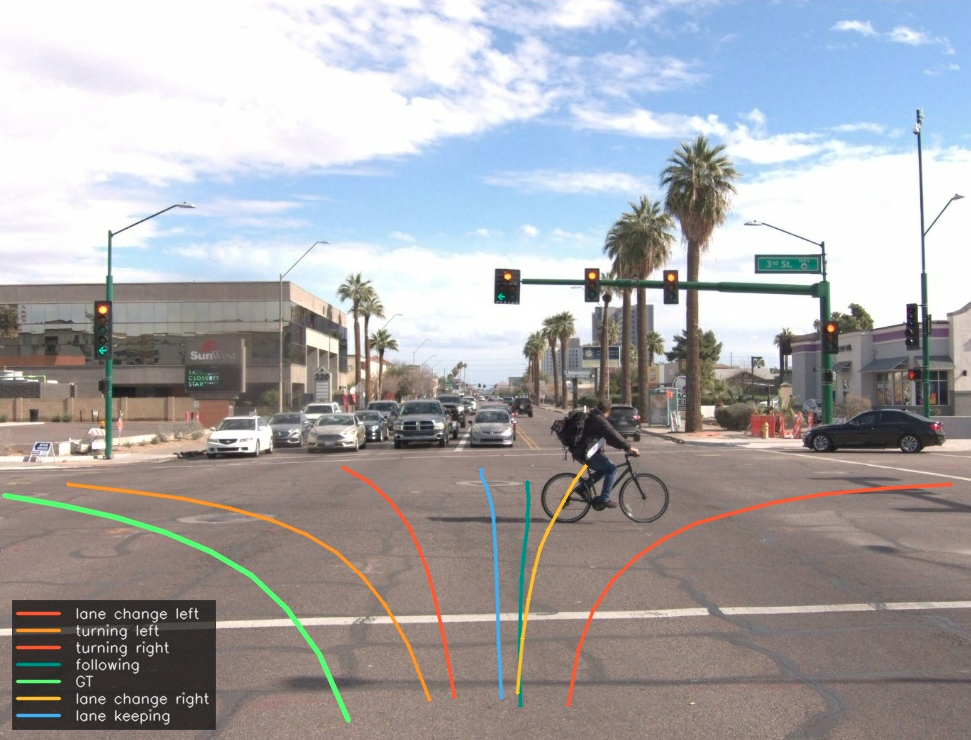} &
    \includegraphics[height=0.27\linewidth,keepaspectratio]%
      {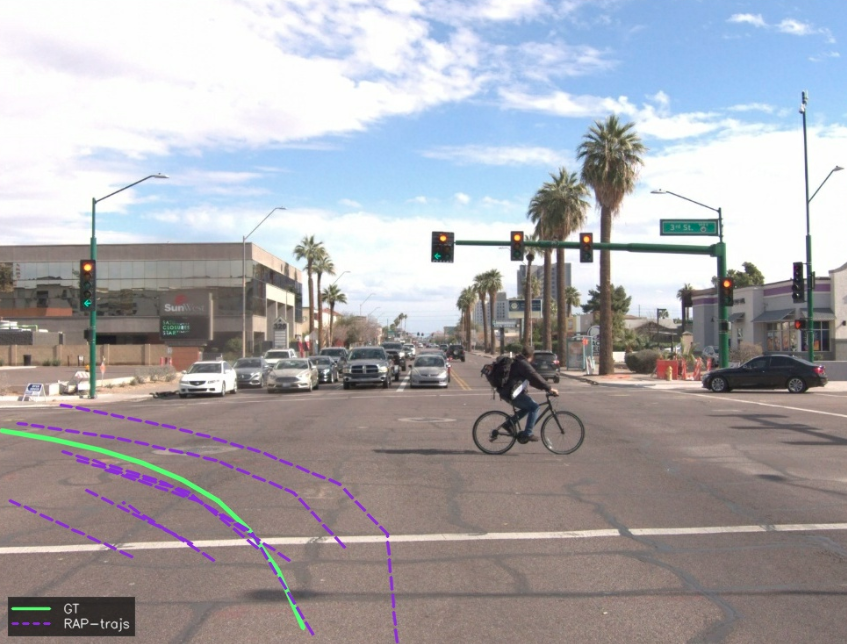} &
    \includegraphics[height=0.27\linewidth,keepaspectratio]%
      {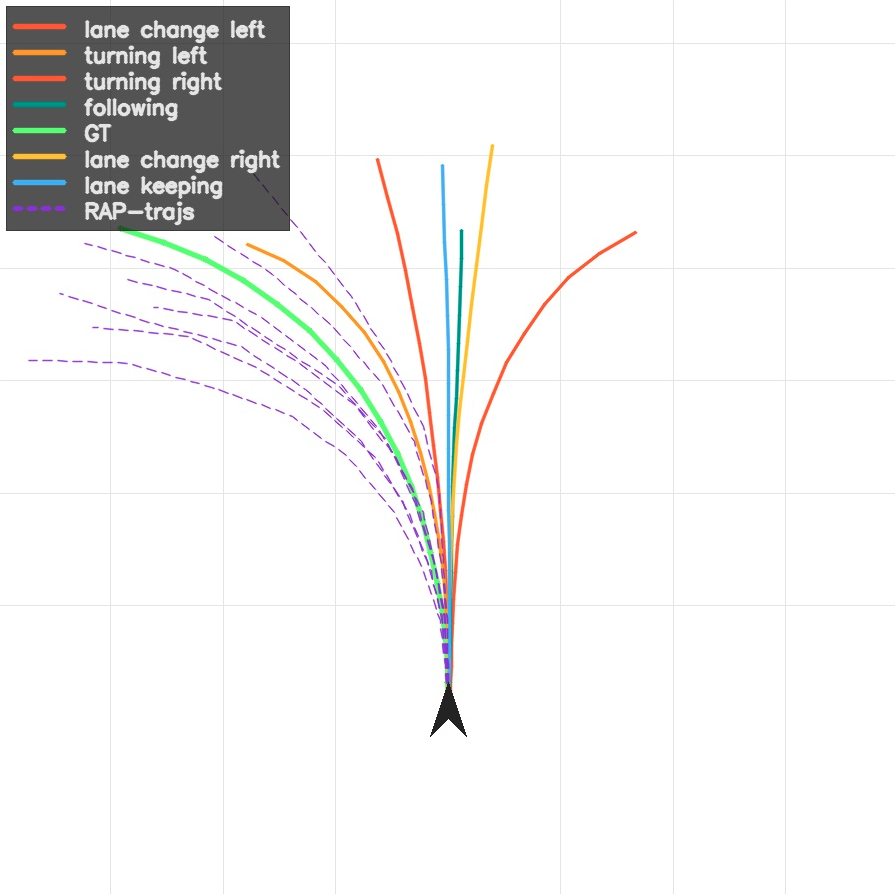} \\
  \end{tabular}%
  }
  \caption{\textbf{Action emergence on long-tail scenes.}
  Across two representative scenes, SI produces intent-faithful trajectory
  families, while RAP~\cite{feng2025rap} collapses to a narrow proposal mode;
  the BEV overlays highlight the contrast.}
  \label{fig:action_emergence}
\end{figure}

\subsection{Action Emergence Demonstration}
\label{subsec:action_emergence}

We probe action emergence qualitatively by selecting scenes whose
semantically appropriate plan is \emph{not} the statistically dominant
(forward-cruising) mode, and comparing two trajectory families on the
same scene: SI's intent-sweep trajectories over a curated subset of
intent classes (one CFG-steered trajectory per selected intent),
against the $8$ proposals emitted by RAP~\cite{feng2025rap}.
We pick RAP as our comparison baseline because it currently sits at
the top of the Waymo E2E Vision-based Driving Challenge
leaderboard~\cite{xu2025wode2e}, making it the strongest publicly
available single-mode end-to-end driving model on this benchmark.
\autoref{fig:action_emergence} arranges two scenes as two rows; each
row places SI (left), RAP (middle), and a joint top-down BEV overlay
of both (right) on the identical scene, since the intent-driven
diversity is most legible from a top-down perspective.
This empirically realizes the per-paradigm schema sketched in
\autoref{fig:diversity_compare}: single-mode generators cannot steer
across intents and collapse onto the dominant prior, whereas SI
delivers the intent-faithful diversity~--~the hallmark of action
emergence~--~that the schema anticipates for our design.

\subsection{Multi-Intent Trajectory Quality}
\label{subsec:multi_intent_quality}

\begin{figure}[t]
  \centering
  \setlength{\tabcolsep}{2pt}
  \renewcommand{\arraystretch}{1.05}
  \noindent\makebox[\linewidth][c]{%
  \begin{tabular}{@{}c@{\hspace{4pt}}c@{\hspace{4pt}}c@{}}
    \includegraphics[width=0.325\linewidth]%
      {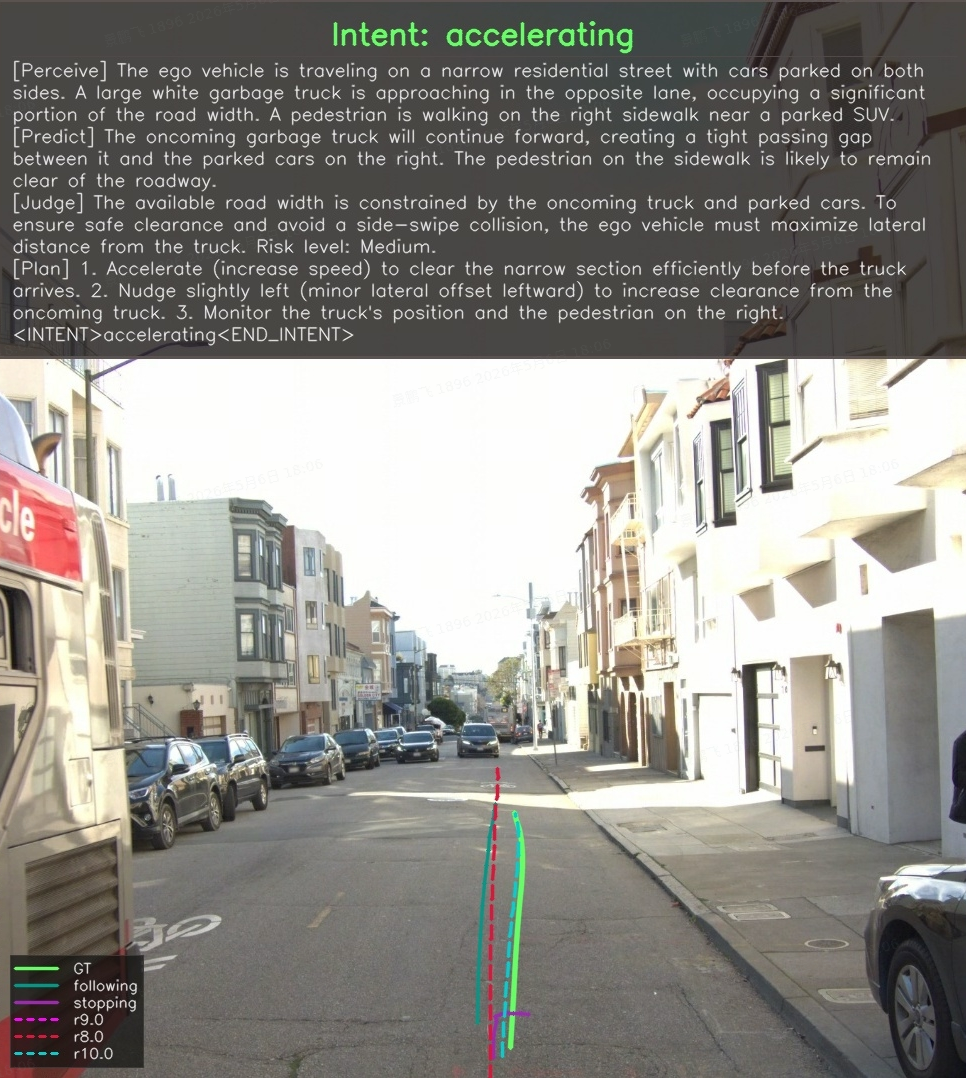} &
    \includegraphics[width=0.325\linewidth]%
      {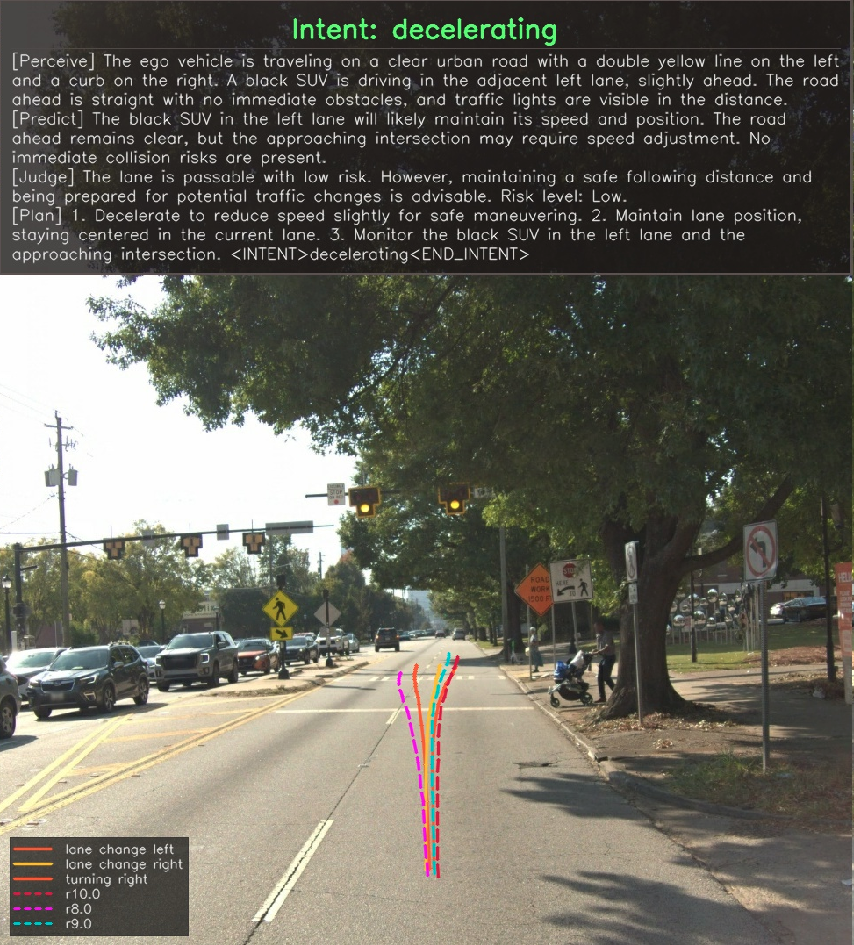} &
    \includegraphics[width=0.325\linewidth]%
      {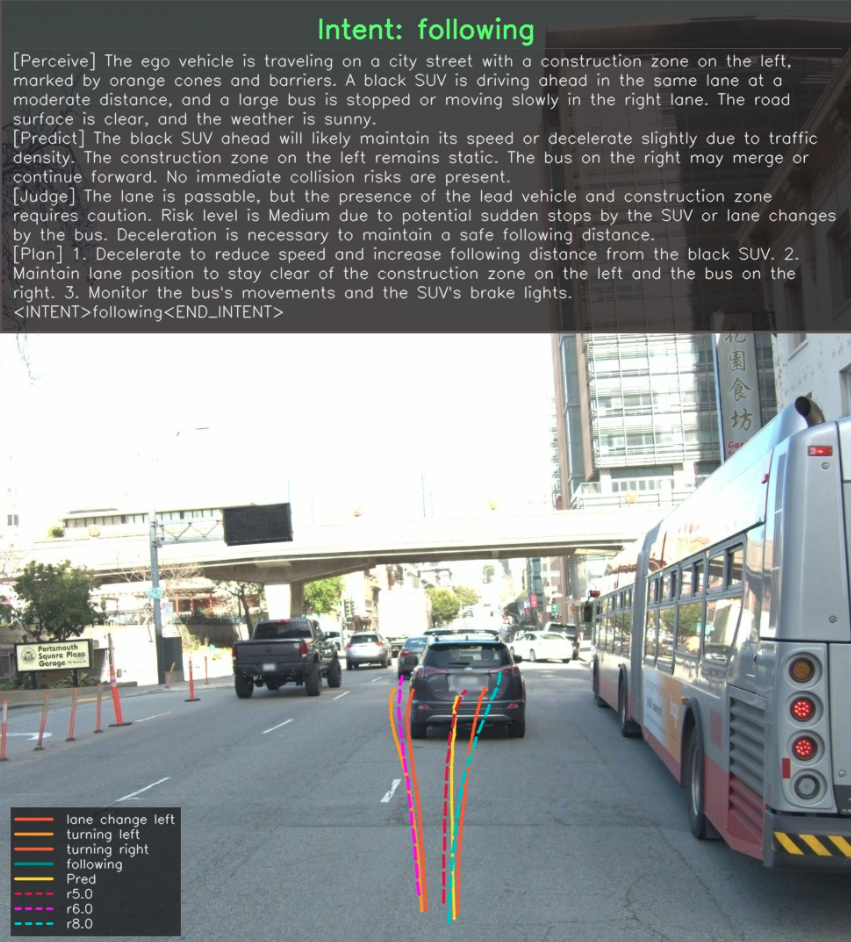} \\
  \end{tabular}%
  }
  \caption{\textbf{Multi-intent trajectory quality on RFS-annotated Waymo E2E scenes.}}
  \label{fig:multi_intent_quality}
\end{figure}

Having demonstrated that SI produces a geometrically diverse family of
intent-conditioned trajectories (\autoref{subsec:action_emergence}), we now
verify that each intent-driven trajectory is \emph{individually} high-quality
rather than an arbitrary deviation away from the dominant mode.  Since each
driving scene in standard benchmarks carries only a \emph{single} GT
trajectory, directly scoring a non-GT intent against the GT would trivially mark
that intent as ``wrong.''  We therefore leverage the Waymo E2E validation
set~\cite{xu2025wode2e}, which annotates each scene with \emph{three}
rater-feedback-score (RFS) trajectories in addition to the GT, each reflecting
a plausible human-preferred maneuver scored by expert raters.  Across the three
validation scenes shown in \autoref{fig:multi_intent_quality}, SI's
multi-intent trajectories align with distinct human-rated RFS alternatives,
indicating high-quality intent-conditioned diversity and demonstrating that the
diversity observed in \autoref{fig:action_emergence} is coverage of the
human-rated action repertoire rather than arbitrary variance.

Beyond single-clip intent-following, SI is designed to maintain
\emph{coherent} intent commitments along a multi-clip driving
horizon via its streaming-memory bank and prev-intent token
(\autoref{subsec:streaming_intent}).  A qualitative demonstration
on a pedestrian-crossroad episode~--~\emph{stopping} $\to$
\emph{waiting} $\to$ \emph{accelerating} $\to$ \emph{cruising}
$\to$ \emph{decelerating}~--~showing SI's per-clip CoT and
trajectory across a $\sim\!1.5\,\mathrm{s}$ window is provided
in~\autoref{app:streaming_consistency}.

\subsection{CFG Distillation for Halving Inference Cost}
\label{subsec:cfg_distillation}

The CFG-guided sampling of Eq.~\eqref{eq:cfg} runs two backbone
forwards per denoising step~--~one conditioned on the decoded intent
$k$ and one on the uncond index $K$~--~doubling the action-head
inference cost.
After the main SI model converges, we freeze every parameter except
a small \emph{DistilledIntentEmbedder} ($\sim\!21\,\mathrm{M}$
trainable out of the $\sim\!2.46\,\mathrm{B}$ backbone) and train it
to reproduce the teacher's CFG-combined velocity field in a
\emph{single} forward pass.
The student embedder has a residual form
$e_{\text{dist}}(k) = \bar{e}(k) + \mathrm{MLP}(\bar{e}(k))$, with
the linear warm-start
$\bar{e}(k) = w\,e(k) - (w{-}1)\,e(K)$~--~the closed-form
CFG-effective vector an idealized linear backbone would
consume~--~and the MLP absorbs the non-linear correction the actual
Qwen3-VL backbone introduces.
Supervision is a per-step velocity-MSE against the CFG teacher,
evaluated on the teacher's own denoising trajectory so teacher and
student stay aligned at every integration step
(\autoref{subsec:streaming_intent}).
At deployment, the FM head is routed through $e_{\text{dist}}(k)$
in place of $e(k)$: the intent-steered velocity field is recovered
in a single backbone forward per denoising step, halving the
action-head inference cost.

\autoref{tab:cfg_distill} reports Waymo E2E RFS val metrics for the
two-pass CFG teacher and the single-pass distilled student on the
same $438$-sample set as \autoref{tab:planning}.
The student trails the teacher by $-0.021$ absolute RFS and $-0.9$
TR points~--~within run-to-run noise on this benchmark~--~while
\emph{improving} ADE/FDE at both $3$\,s and $5$\,s horizons.
Distillation therefore halves inference cost with no meaningful
degradation of planning quality.

\begin{table}[t]
  \centering
  \small
  \setlength{\tabcolsep}{5pt}
  \caption{\textbf{CFG teacher vs.\ distilled student on Waymo E2E
  RFS val ($438$ samples).}
  The distilled single-pass student preserves the teacher's
  planning quality while halving the action-head inference cost.
  $\uparrow$ / $\downarrow$: higher / lower is better.}
  \label{tab:cfg_distill}
  \begin{tabular}{l c c c c c c c}
    \toprule
    \multirow{2}{*}{Model}
      & \multirow{2}{*}{Cost}
      & \multirow{2}{*}{RFS $\uparrow$}
      & \multirow{2}{*}{TR (\%) $\uparrow$}
      & \multicolumn{2}{c}{ADE (m) $\downarrow$}
      & \multicolumn{2}{c}{FDE (m) $\downarrow$} \\
    \cmidrule(lr){5-6}\cmidrule(lr){7-8}
      & & & & 3\,s & 5\,s & 3\,s & 5\,s \\
    \midrule
    CFG teacher (2-pass)       & $2{\times}$
      & 7.898          & 61.4           & 1.123          & 2.461          & 2.600          & 6.068          \\
    Distilled student (1-pass) & $\mathbf{1{\times}}$
      & 7.877          & 60.5           & \textbf{1.112} & \textbf{2.423} & \textbf{2.563} & \textbf{5.958} \\
    \bottomrule
  \end{tabular}
\end{table}

\section{Related Work}
\label{sec:related}

End-to-end trajectory planners broadly split along two axes.
The first direction formulates planning as direct trajectory
regression, planning-token scoring, or autoregressive
language--action generation~\citep{hu2023planning,chen2024vadv2,zhou2025autovla,rowe2025poutine,luo2025adathinkdrive,chen2026curiousvla},
with more recent work modelling the multi-modal future via
diffusion or flow matching~\citep{liao2025diffusiondrive,xing2025goalflow,xu2025wamflow,li2025recogdrive}.
The second direction avoids unconstrained continuous generation
by pre-banking a trajectory vocabulary or by generating multiple
candidates and selecting one afterward~\citep{chai2020multipath,phanminh2020covernet,chen2024vadv2,sun2026sparsedrivev2,gao2026rad2}.
Both lines advance trajectory feasibility, diversity, or
inference efficiency, but neither explicitly binds a reasoned
discrete driving \emph{intent} to the trajectory generator:
generative methods rely on data priors / goal points / route
commands, while candidate-scoring methods pick from a bank or
rerank candidates via a learned discriminator.  To our knowledge,
SI is the first end-to-end VLA to demonstrate intent-faithful
trajectory following from a single trained model, without any
pre-built trajectory bank or hand-tuned post-hoc selector.
A detailed per-method discussion, including how each prior line
relates to the intent-faithful controllability property we
target, is given in~\autoref{app:related_work}.

\section{Discussion and Conclusion}
\label{sec:discussion_conclusion}

\textbf{Limitations and future directions.}
Streaming Intent is a purely data-driven mechanism, so its behavior is directly
shaped by the quality and coverage of the intent labels used during training.
In our current setup, some minority intents~--~most notably \emph{reversing}
and \emph{parking}~--~remain under-represented in forward-driving corpora, and
SI's trajectory fidelity on these classes trails better-covered intents
accordingly. Scaling the annotation pipeline to larger driving corpora,
e.g.\ NVIDIA's Alpamayo dataset~\citep{wang2025alpamayo}, and enriching the
intent taxonomy with balanced CoT-grounded labels are therefore natural next
steps. A second limitation is evaluation: while
\autoref{subsec:multi_intent_quality} qualitatively shows that SI's
intent-conditioned trajectories align with both the single ground-truth
trajectory and human-rated RFS alternatives, existing metrics such as ADE, FDE,
TR, and RFS score only one predicted plan against one reference and do not
naturally evaluate multiple simultaneous, intent-distinct plans for the same
scene. Future work should develop quantitative protocols that jointly measure
per-intent kinematic plausibility, alignment with rater-preferred alternatives,
and mode coverage.

\textbf{Conclusion.}
We introduced \emph{action emergence} as a target capability for end-to-end
autonomous driving and argued that existing autoregressive, diffusion /
flow-matching, and VLA-based planners lack the key property needed to approach
it: \emph{intent-faithful controllability}. We proposed \textbf{Streaming
Intent}, which grounds driving intent through four-step chain-of-thought
reasoning and carries intent commitments across clips with a prev-intent memory
token. Instantiated as \textbf{SI}, our model uses an AR-decoded intent token to
guide a shared flow-matching action head via classifier-free guidance, producing
controllable trajectory families from a single end-to-end-trained VLA. On Waymo
End-to-End, SI achieves competitive planning performance while demonstrating
intent-faithful controllability without a pre-built trajectory bank or
hand-coded post-hoc selector, suggesting a practical path toward action
emergence through structurally aligned language--action learning.

\bibliographystyle{abbrvnat}
\bibliography{refs}

\newpage

\appendix


\section{Extended Related Work}
\label{app:related_work}

This section expands the one-paragraph related-work summary
in~\autoref{sec:related} with per-method discussion, organised by
the two axes identified there (generative trajectory models vs.\
pre-banked vocabulary / post-hoc selection).

\paragraph{Autoregressive and generative trajectory models.}
End-to-end trajectory generation for autonomous driving has recently evolved
along two complementary directions.  The first direction formulates planning as
direct trajectory prediction, planning-token scoring, or autoregressive
language--action generation.  Planning-oriented systems such as
UniAD~\citep{hu2023planning} integrate perception, prediction, occupancy
forecasting, and planning through unified query interfaces, while
VADv2~\citep{chen2024vadv2} moves beyond deterministic regression by
discretizing the continuous planning action space into a planning vocabulary and
predicting a scene-conditioned probability distribution over planning actions.
More recent VLA-based planners further cast driving as language--action
generation: AutoVLA~\citep{zhou2025autovla} tokenizes continuous trajectories
into physical action tokens and generates reasoning and actions with a
vision--language--action model; Poutine~\citep{rowe2025poutine} performs
vision--language--trajectory pre-training followed by reinforcement-learning
post-training; AdaThinkDrive~\citep{luo2025adathinkdrive} introduces adaptive
fast/slow reasoning to decide when chain-of-thought is necessary; and
CuriousVLA~\citep{chen2026curiousvla} improves exploration in driving VLA
models by addressing narrow-policy behavior.  The second direction models the
multimodal distribution of future trajectories with diffusion or flow matching.
DiffusionDrive~\citep{liao2025diffusiondrive} introduces a truncated diffusion
model for efficient end-to-end planning; GoalFlow~\citep{xing2025goalflow} uses
goal-point guidance and flow matching to generate multimodal trajectories;
WAM-Flow~\citep{xu2025wamflow} applies discrete flow matching over structured
trajectory tokens for parallel coarse-to-fine planning; and
ReCogDrive~\citep{li2025recogdrive} combines cognitive VLA reasoning with a
diffusion-based planning framework.  These methods improve trajectory diversity,
feasibility, reasoning ability, or inference efficiency, but their generated
plans are still primarily governed by data priors, route commands, goal points,
sampled candidates, or scalar planning rewards.  They do not explicitly bind a
reasoned discrete driving intent to the final trajectory generation process,
and therefore do not provide intent-faithful controllability in which changing
the committed intent produces a correspondingly changed, semantically
consistent trajectory for the same scene.

\paragraph{Pre-banked trajectory vocabularies and post-hoc selection.}
Another line of work avoids unconstrained continuous generation by pre-banking a
trajectory vocabulary or by generating multiple candidates and selecting one
afterward.  Classical multimodal prediction methods such as
MultiPath~\citep{chai2020multipath} and CoverNet~\citep{phanminh2020covernet}
already demonstrated the effectiveness of anchor or trajectory-set
representations, and recent end-to-end planners extend this idea to ego
planning.  VADv2~\citep{chen2024vadv2} discretizes the continuous planning
action space into a large planning vocabulary and learns a probabilistic
distribution over candidate planning actions.  SparseDriveV2~\citep{sun2026sparsedrivev2}
pushes this scoring paradigm further by factorizing a dense trajectory
vocabulary into geometric paths and velocity profiles, then performing coarse
factorized scoring followed by fine-grained scoring over a small set of
composed candidates.  Post-hoc selection methods take the same idea to
generative planners: RAD-2~\citep{gao2026rad2} uses a diffusion-based generator
to produce diverse trajectory candidates and an RL-optimized discriminator to
rerank them according to long-term driving quality.  While effective on
benchmark metrics, these methods treat planning as candidate coverage plus
score maximization; the selected trajectory is the one preferred by a
vocabulary scorer, discriminator, or reward model, rather than the one causally
determined by an explicitly reasoned intent.  Consequently, they cannot
guarantee that a requested or inferred intent---for example turning, yielding,
accelerating, decelerating, or nudging---will be faithfully reflected in the
geometry and speed profile of the executed trajectory.  In contrast, our
Streaming Intent mechanism grounds intent through semantic chain-of-thought,
carries it temporally across clips, and uses the decoded intent to guide the
shared flow-matching action head.  To our knowledge, SI is the first
end-to-end VLA model to demonstrate intent-faithful trajectory following
without relying on a pre-built trajectory bank or a hand-tuned post-hoc
trajectory selector.

\section{Data Construction Details}
\label{app:data_construction}

This appendix expands on the four-stage streaming data annotation
pipeline of~\autoref{subsec:data_construction}.
Sec.~\ref{app:taxonomy} documents the driving-intent and meta-action
taxonomies together with their design rationale;
Sec.~\ref{app:rules} gives the exact kinematic rules that derive
meta-actions from the ego trajectory;
Sec.~\ref{app:prompt} shows the VLM prompt used to relabel intent
from video plus kinematic evidence;
and Sec.~\ref{app:worked_example} walks through one concrete
training sample end-to-end.

\subsection{Intent and Meta-Action Taxonomies}
\label{app:taxonomy}

\paragraph{20-class driving intent.}
Our intent space is a closed 20-class taxonomy, grouped into four
semantic clusters to cover the common driving maneuvers a human
rater would name when asked ``what is the ego vehicle doing?''
The full list is given in~\autoref{tab:app_intent_taxonomy}.
The design rationale is: (i)~cover all \emph{longitudinal}
primitives a driver commits to (steady-state, transient
accel/decel, starting, stopping, waiting, car-following);
(ii)~cover the standard \emph{lateral} maneuvers (lane keeping
vs.\ lane change, turning at intersections, U-turn);
(iii)~carve out the few \emph{complex-context} maneuvers whose
kinematic signature is ambiguous without visual context
(yielding, merging, overtaking, obstacle avoidance);
(iv)~keep the taxonomy \emph{closed} and \emph{mutually
exclusive} so that intent supervision is unambiguous and the
flow-matching head's CFG embedder has a compact index space.

\begin{table}[h]
  \centering
  \small
  \setlength{\tabcolsep}{6pt}
  \caption{20-class driving-intent taxonomy, grouped by semantic
  category.}
  \label{tab:app_intent_taxonomy}
  \resizebox{0.95\linewidth}{!}{%
  \begin{tabular}{l p{0.76\linewidth}}
    \toprule
    Group & Classes (idx in parentheses) \\
    \midrule
    Longitudinal (straight-line, no sustained yaw) &
      \texttt{cruising} (0), \texttt{lane\_keeping} (1),
      \texttt{following} (2), \texttt{starting} (8),
      \texttt{stopping} (9), \texttt{waiting} (10),
      \texttt{accelerating} (11), \texttt{decelerating} (12),
      \texttt{braking} (13) \\
    Lateral (direction or lane change) &
      \texttt{lane\_change\_left} (3),
      \texttt{lane\_change\_right} (4),
      \texttt{turning\_left} (5), \texttt{turning\_right} (6),
      \texttt{u\_turn} (7) \\
    Complex context-dependent &
      \texttt{yielding} (14), \texttt{overtaking} (15),
      \texttt{merging} (16), \texttt{avoiding\_obstacle} (17) \\
    Special low-speed &
      \texttt{parking} (18), \texttt{reversing} (19) \\
    \bottomrule
  \end{tabular}%
  }
\end{table}

\paragraph{Meta-actions: $7$ longitudinal + $7$ lateral.}
Meta-actions are a deterministic, rule-based discretization of the
ego's future $3\,\mathrm{s}$ kinematic window, computed with zero
VLM involvement.
They serve two roles downstream: they are an explicit input hint to
the VLM at Stage~3, and they are the kinematic consistency
validator that rejects hallucinated intent labels.
The two axes:
\begin{itemize}[leftmargin=1.4em,itemsep=0pt,topsep=2pt]
  \item \textbf{Longitudinal (7):}
    \texttt{stop}, \texttt{reverse}, \texttt{stopping},
    \texttt{starting}, \texttt{accelerate}, \texttt{decelerate},
    \texttt{maintain\_speed}.
  \item \textbf{Lateral (7):}
    \texttt{steer\_left}, \texttt{steer\_right},
    \texttt{nudge\_left}, \texttt{nudge\_right},
    \texttt{reverse\_left}, \texttt{reverse\_right},
    \texttt{maintain}.
\end{itemize}
The $7{+}7$ space is the minimum resolution that covers every
intent-disambiguating kinematic pattern: stopping vs.\ starting
are direction-of-motion events; \texttt{accelerate} /
\texttt{decelerate} / \texttt{maintain\_speed} separate the three
continuous-motion regimes; and on the lateral axis the ``steer''
vs.\ ``nudge'' distinction separates intersection-scale turns
($\sim$5$\,^\circ$ of yaw, several metres of lateral offset) from
lane-change-scale manoeuvres ($\sim$1$\,^\circ$, decimetre-scale
offset), which is the exact call that rule-based meta-actions can
make but a VLM cannot, and vice versa.

\subsection{Rule-Based Meta-Action Derivation}
\label{app:rules}

The meta-action labeler reads the ego's $3\,\mathrm{s}$ future
window (speed, heading, BEV offset) and classifies it into one of
the $7{+}7$ classes with the thresholds
in~\autoref{tab:app_meta_thresholds}.
\emph{Longitudinal} is decided first, then \emph{lateral}.

\begin{table}[h]
  \centering
  \small
  \setlength{\tabcolsep}{6pt}
  \caption{Thresholds used by the deterministic meta-action labeler.
  ``Sustained yaw'' fires when at least $80\%$ of the per-frame
  yaw rates over the $3\,\mathrm{s}$ window share sign \emph{and}
  their absolute mean exceeds $1\,^\circ/\mathrm{frame}$.}
  \label{tab:app_meta_thresholds}
  \begin{tabular}{l l}
    \toprule
    Symbol & Value / meaning \\
    \midrule
    \texttt{SPEED\_STOP}           & $0.3\,\mathrm{m/s}$ (below $\Rightarrow$ vehicle is stationary) \\
    \texttt{LON\_SPEED\_RATIO}     & $0.15$ (fractional speed delta threshold) \\
    \texttt{LON\_MIN\_DELTA}       & $0.5\,\mathrm{m/s}$ (absolute speed delta floor) \\
    \texttt{REVERSE\_DIST}         & $0.5\,\mathrm{m}$ (backward distance $\Rightarrow$ reverse) \\
    \texttt{LAT\_OFFSET\_STEER}    & $1.5\,\mathrm{m}$ (lateral offset for \texttt{steer\_*}) \\
    \texttt{LAT\_OFFSET\_NUDGE}    & $0.3\,\mathrm{m}$ (lateral offset for \texttt{nudge\_*}) \\
    \texttt{YAW\_CHANGE\_STEER}    & $5.0\,^\circ$ over window (threshold for \texttt{steer\_*}) \\
    \texttt{YAW\_CHANGE\_NUDGE}    & $1.5\,^\circ$ over window (threshold for \texttt{nudge\_*}) \\
    \texttt{SUSTAINED\_YAW\_RATIO} & $0.8$ (fraction of frames sharing yaw sign) \\
    \bottomrule
  \end{tabular}
\end{table}

\paragraph{Longitudinal decision.} Let $v_s, v_e, v_{\max}$ be the
start, end, and max speeds in the window, and let $\Delta x$ be
forward displacement.
The classifier returns, in order:
\texttt{stop} if $v_{\max} \le$ \texttt{SPEED\_STOP};
\texttt{reverse} if $\Delta x < -$\texttt{REVERSE\_DIST};
\texttt{stopping} if $v_s >$ \texttt{SPEED\_STOP} and $v_e \le$
\texttt{SPEED\_STOP};
\texttt{starting} if $v_s \le$ \texttt{SPEED\_STOP} and $v_e >$
\texttt{SPEED\_STOP};
\texttt{accelerate} / \texttt{decelerate} if
$|v_e - v_s| > \max($\texttt{LON\_MIN\_DELTA},
$v_s \cdot$\texttt{LON\_SPEED\_RATIO}$)$ with the appropriate sign;
otherwise \texttt{maintain\_speed}.

\paragraph{Lateral decision.} Given sustained-yaw direction $d$,
total lateral offset $\Delta y$ (positive $=$ left), and total
heading change $|\Delta\theta|$:
\texttt{steer\_$d$} if $d$ is defined;
\texttt{steer\_left}/\texttt{steer\_right} if
$|\Delta y| >$ \texttt{LAT\_OFFSET\_STEER} and $|\Delta\theta| >$
\texttt{YAW\_CHANGE\_STEER};
\texttt{nudge\_left}/\texttt{nudge\_right} if
$|\Delta y| >$ \texttt{LAT\_OFFSET\_NUDGE} and $|\Delta\theta| >$
\texttt{YAW\_CHANGE\_NUDGE};
\texttt{reverse\_left}/\texttt{reverse\_right} under the reverse
branch; otherwise \texttt{maintain}.

\subsection{VLM Intent Annotation Prompt}
\label{app:prompt}

Stage~3 runs a single Qwen3.5-Plus forward per clip with the
structured prompt below.
The prompt injects four evidence blocks~--~kinematic facts, the
4-step CoT decoded earlier in the same stage, the rule
pre-classification, and the closed 20-class taxonomy~--~and
requires the VLM to emit exactly one JSON object whose
\texttt{intent} field is a taxonomy key.
Hard constraints forbid the VLM from disagreeing with either
(a)~the CoT Plan step (which is itself hallucination-filtered
upstream) or (b)~the meta-action axes.

\begin{figure}[h]
  \centering
  \footnotesize
\begin{tabular}{@{}p{0.96\linewidth}@{}}
  \toprule
  \textbf{Fresh-label prompt (abridged).} \\
  \midrule
  \ttfamily
TASK: Driving intent classification (20-class taxonomy).

You are given sampled front-camera frames of a $\sim$9\,s clip
(4\,s past + 5\,s future, in chronological order) plus kinematic
ground-truth and a 4-step chain-of-thought.  Assign exactly ONE
intent label from the taxonomy below.

HARD CONSTRAINTS (violations are rejected):

1. Your intent MUST be logically coherent with the 4-step CoT
   Plan step.  If Plan says ``slow down and yield'', do not
   output \textit{accelerating} or \textit{turning\_left}.

2. Your intent MUST NOT contradict the kinematic facts.
   E.g.\ if lateral\_action $=$ \textit{steer\_left}, do not
   output \textit{turning\_right} / \textit{lane\_change\_right};
   if longitudinal\_action $=$ \textit{stop}, do not output
   \textit{cruising}.

3. Use the VIDEO only to disambiguate what kinematics + CoT
   cannot decide, e.g.\ \textit{steer\_left} at an intersection
   $\Rightarrow$ \textit{turning\_left};
   \textit{nudge\_right} around a parked car $\Rightarrow$
   \textit{avoiding\_obstacle}, not \textit{lane\_change\_right}.

\rmfamily \\
  \{4-STEP CHAIN-OF-THOUGHT: Perceive / Predict / Judge / Plan\} \\
  \{KINEMATIC FACTS: lon / lat action, start/end speed,
  heading change, total distance, lateral displacement\} \\
  \{RULE PRE-CLASSIFICATION: candidate intent, reason, needs\_llm\} \\
  \{INTENT TAXONOMY: full 20-class list with descriptions,
  grouped by longitudinal / lateral / complex / special\} \\

  \ttfamily
Respond with a single JSON object:

\{"intent": "$\langle$one of the keys$\rangle$",
  "confidence": "$\langle$high$|$medium$|$low$\rangle$",
  "rationale": "$\langle$one sentence citing CoT Plan + video$\rangle$",
  "coherent\_with\_cot": true\}
\rmfamily \\
  \bottomrule
\end{tabular}
  \caption{The prompt used to produce one intent label per clip in
  Stage~3 of~\autoref{subsec:data_construction}.  A parallel
  ``revise'' prompt (kept-biased) is run whenever a prior intent
  label needs adjudication; it shares the four evidence blocks and
  only adds a prior-label note with a ``keep by default, revise on
  hard contradiction'' decision policy.}
  \label{fig:app_intent_prompt}
\end{figure}

\subsection{A Worked Example}
\label{app:worked_example}

\autoref{tab:app_worked_example} walks through one training clip
(seq~\texttt{0bec10cd\ldots}, segment
\texttt{front-3\_f000149\_f000188\_4s}) end-to-end: the front-3
stitched camera view, the rule-derived meta-action, the
VLM-adjudicated intent, the full 4-step CoT, and a sample of the
auxiliary VQA pairs produced by the same Stage-3 VLM call.
Together these constitute one entry in the streaming training set.

\begin{table}[h]
  \centering
  \small
  \setlength{\tabcolsep}{6pt}
  \renewcommand{\arraystretch}{1.15}
  \caption{Worked example: one training clip with its front-3
  view, meta-action + intent, 4-step CoT, and auxiliary VQAs.}
  \label{tab:app_worked_example}
  \begin{tabular}{@{}p{0.14\linewidth} p{0.80\linewidth}@{}}
    \toprule
    \multicolumn{2}{c}{%
      \includegraphics[width=0.95\linewidth]%
        {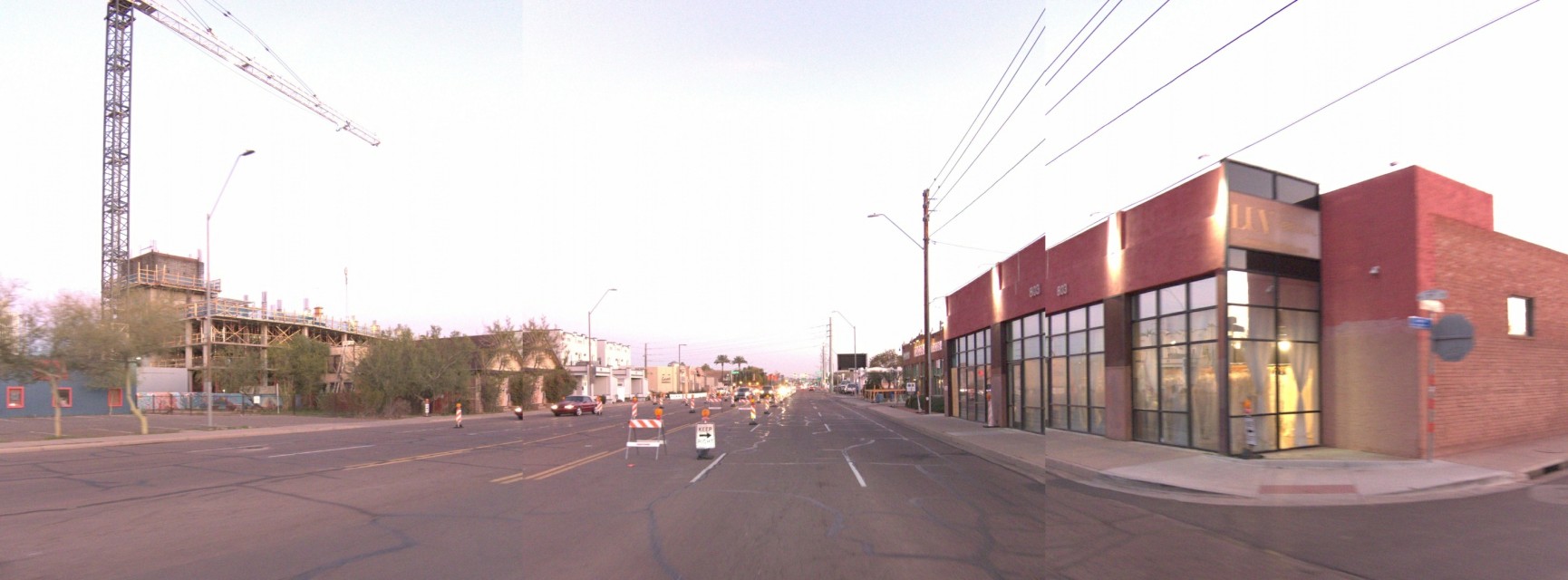}%
    } \\
    \multicolumn{2}{c}{\footnotesize \emph{Front-3 stitched camera view of the clip.}} \\
    \midrule
    Meta-action &
    Longitudinal: \texttt{maintain\_speed};
    lateral: \texttt{nudge\_right} \\
    Intent &
    \texttt{avoiding\_obstacle} (VLM-adjudicated; rule pre-class
    was \texttt{lane\_change\_right}, overridden after video
    disambiguation) \\
    \midrule
    \multirow{4}{=}{4-step CoT} &
    \textsc{Perceive}: Ego travels on a multi-lane urban road
    with a construction zone on the left (orange cones, barriers,
    a large crane); the right lane is clear; traffic ahead is
    light. \\
    &
    \textsc{Predict}: The construction zone remains static with no
    encroachment into the ego lane; road ahead is clear; no
    pedestrians / cyclists on the right sidewalk. \\
    &
    \textsc{Judge}: Current lane is passable; a slight rightward
    nudge maximizes clearance from the construction zone; risk
    low; maintaining speed is safe. \\
    &
    \textsc{Plan}: 1.~Maintain speed; 2.~nudge slightly right to
    increase clearance from the left construction zone;
    3.~monitor the construction barriers and distant traffic.
    $\to\,\langle\textsc{INTENT}\rangle$\,\texttt{avoiding\_obstacle}. \\
    \midrule
    \multirow{4}{=}{VQA samples} &
    \textbf{Q} \emph{(existence)}: Is there a sedan visible in the
    center lane ahead?~~\textbf{A}: No~--~the center lane is
    empty; only distant cones and a distant red car in the left
    lane. \\
    &
    \textbf{Q} \emph{(spatial\_ref)}: Where is the barrier in the
    center lane ahead relative to the ego vehicle?~~\textbf{A}:
    $\sim\!25\,\mathrm{m}$ ahead; orange-and-white striped
    traffic barrier with a ``KEEP RIGHT'' sign. \\
    &
    \textbf{Q} \emph{(counting)}: How many infrastructure elements
    are visible near the center lane ahead and right
    roadside?~~\textbf{A}: Three~--~a ``KEEP RIGHT'' barrier, an
    orange traffic cone, and a circular traffic sign on the right. \\
    &
    \textbf{Q} \emph{(occlusion)}: Is the barrier in the center
    lane ahead partially or fully occluded?~~\textbf{A}: No~--~it
    is clearly visible with no vehicles, pedestrians, or
    structures blocking its view. \\
    \bottomrule
  \end{tabular}
\end{table}

\section{Streaming Intent Consistency}
\label{app:streaming_consistency}

\begin{figure}[tp]
  \centering
  \setlength{\tabcolsep}{0pt}
  \renewcommand{\arraystretch}{0.9}
  \begin{tabular}{@{}c@{}}
    \includegraphics[width=0.78\linewidth]%
      {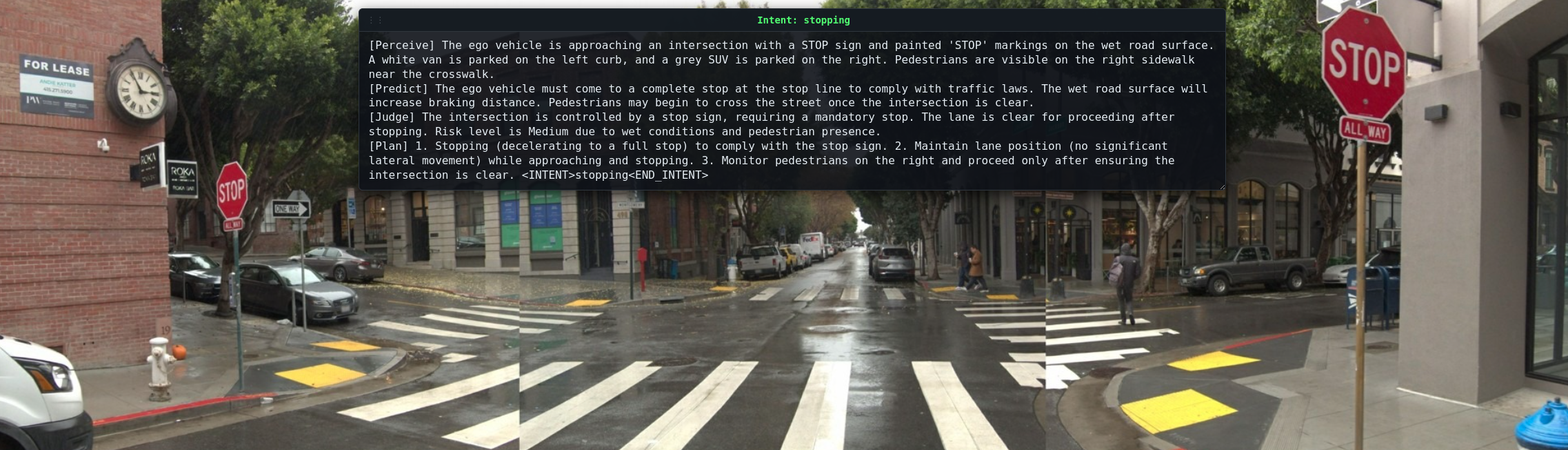} \\
    \includegraphics[width=0.78\linewidth]%
      {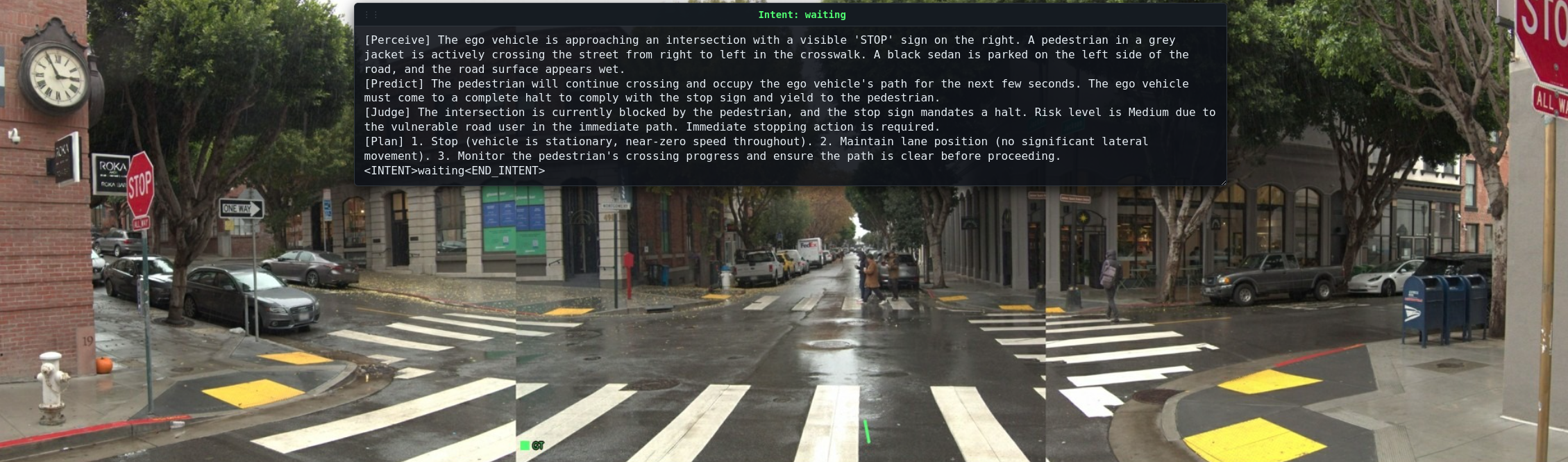} \\
    \includegraphics[width=0.78\linewidth]%
      {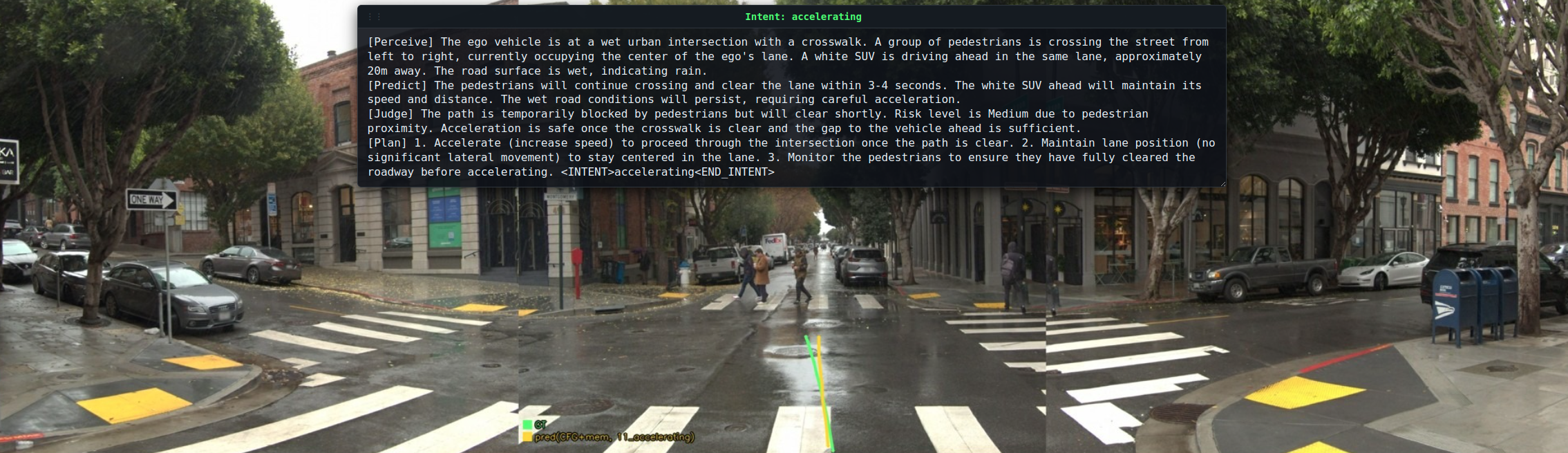} \\
    \includegraphics[width=0.78\linewidth]%
      {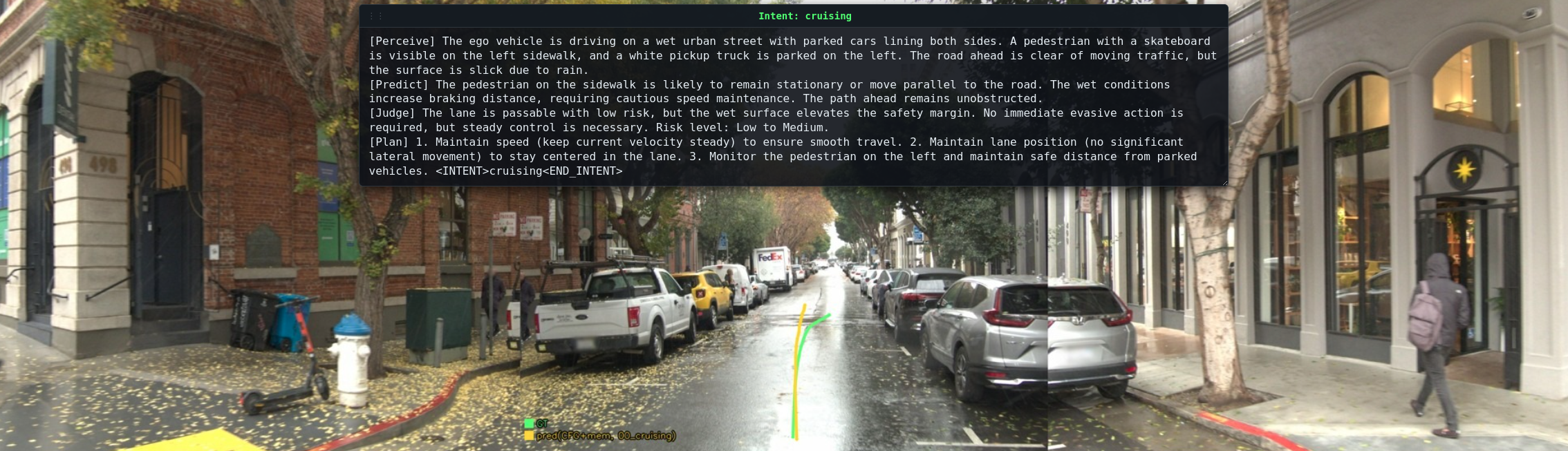} \\
    \includegraphics[width=0.78\linewidth]%
      {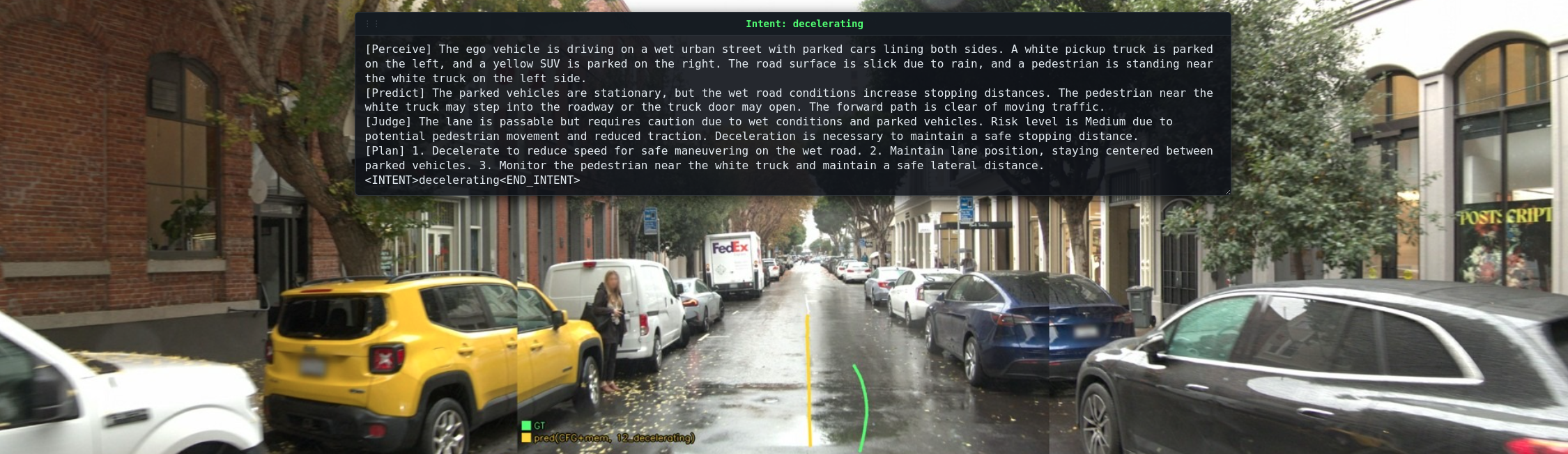} \\
  \end{tabular}
  \caption{\textbf{Streaming Intent consistency on a multi-clip
  pedestrian-crossroad episode.}
  Five per-clip snapshots at
  $t{=}0.13 / 0.43 / 0.63 / 1.18 / 1.43\,\mathrm{s}$
  (top-to-bottom).
  Each panel shows SI's 4-step CoT and decoded intent above the
  front-3 view, with the predicted trajectory overlaid against
  the GT.  Per-clip intent sequence and analysis in text.}
  \label{fig:streaming_consistency}
\end{figure}

This appendix expands the pointer at the end
of~\autoref{subsec:multi_intent_quality}: where the main-text
subsections probe SI at the single-clip level, here we trace SI's
\emph{Streaming Intent} mechanism
(\autoref{subsec:streaming_intent}) across a multi-clip horizon on
which per-clip intent commitments must stay coherent as the scene
evolves.
\autoref{fig:streaming_consistency} shows SI's per-clip decoded
reasoning and trajectory across a $\sim\!1.5\,\mathrm{s}$ window in
which the ego vehicle negotiates a pedestrian-occupied crossroad.
Above each panel the full 4-step CoT reasoning and emitted intent
class are printed; below, the panel shows the predicted trajectory
against the GT at that timestamp.
The per-clip intents evolve smoothly with the scene:
$t{=}0.13\,\mathrm{s}$, the ego encounters the crossroad and
\emph{stops};
$t{=}0.43\,\mathrm{s}$, it continues \emph{waiting} as pedestrians
are crossing;
$t{=}0.63\,\mathrm{s}$, with pedestrians about to clear, it begins
\emph{accelerating};
$t{=}1.18\,\mathrm{s}$, the crossroad is behind the ego and it
settles into \emph{cruising};
$t{=}1.43\,\mathrm{s}$, a new pedestrian appears on the road and
the model switches to \emph{decelerating}.
Each transition is a causal continuation of the preceding CoT
conditioned on the streaming-memory bank and the prev-intent
token~--~no transition requires recomputing the scene from
scratch, and no adjacent clips disagree on the current
commitment.
This is the behavioural signature of Streaming Intent: a single
trained model carries a coherent intent commitment through the
episode, and updates it precisely when the scene warrants.

\end{document}